\documentclass[dvipsnames]{article} % For LaTeX2e
\usepackage{iclr2021_conference,times}
\usepackage{amsmath,amsfonts,bm}

\def\Figref#1{Figure~\ref{#1}}

\def\eqref#1{equation~\ref{#1}}
\def\1{\bm{1}}

\def\1{\bm{1}}
\newcommand{\D}{\mathcal{D}}
\newcommand{\X}{\mathcal{X}}
\newcommand{\Y}{\mathcal{Y}}
\newcommand{\T}{\mathcal{T}}
\newcommand{\K}{\mathcal{K}}

\newcommand{\sws}{\sigma_\omega^2}

\newcommand{\sbs}{\sigma_b^2}

\newcommand{\pp}[1]{\left( #1 \right)}

\newcommand{\mc}{\mathcal}

\newcommand{\infntk}{\Theta}
\newcommand{\infnngp}{\mathcal K}

\newcommand{\tpoint}{x}

\newcommand{\GP}{\mc {GP}}
\newcommand{\e}{{\varepsilon}}

\renewcommand{\P}{\mathbb{P}}

\newcommand{\deq}{\mathrel{\mathop:}=} % aligned define equals
\DeclareMathOperator{\erf}{erf}

\DeclareMathOperator{\relu}{ReLU}
\newcommand*\diff{\mathop{}\!\mathrm{d}}

\DeclareMathAlphabet{\mathsfit}{\encodingdefault}{\sfdefault}{m}{sl}
\SetMathAlphabet{\mathsfit}{bold}{\encodingdefault}{\sfdefault}{bx}{n}

\newcommand{\R}{\mathbb{R}}

\newcommand{\softmax}{\mathrm{softmax}}

\DeclareMathOperator*{\argmax}{arg\,max}

\newcommand{\eq}[1]{\begin{equation}#1\end{equation}}

\newcommand{\al}[1]{\begin{align}#1\end{align}}

\newcommand{\h}[1]{\{{#1}\}}

\usepackage{hyperref}
\usepackage{url}
\usepackage{graphicx}
\usepackage{booktabs}
\usepackage{caption}
\usepackage{subcaption}
\usepackage{enumitem}
\usepackage{siunitx}
\usepackage[utf8]{inputenc} % allow utf-8 input
\usepackage[T1]{fontenc}    % use 8-bit T1 fonts
\usepackage{hyperref}       % hyperlinks
\usepackage{url}            % simple URL typesetting
\usepackage{amsfonts}       % blackboard math symbols
\usepackage{nicefrac}       % compact symbols for 1/2, etc.
\usepackage{microtype}      % microtypography
\usepackage{hyperref}

\usepackage{dcolumn}
\usepackage{multirow}

\usepackage[compact]{titlesec}

\definecolor{darkgreen}{rgb}{0,0.6,0}
\newif\ifshowcomments
\showcommentsfalse

\ifshowcomments
    \usepackage{showlabels}
    \newcommand{\jl}[1]{{\color{red}[JL: #1]}}
    \newcommand{\xlc}[1]{{\color{blue}[XLC: #1]}}
    \newcommand{\ba}[1]{{\color{darkgreen}[BA: #1]}}
    \newcommand{\jp}[1]{{\color{purple}[jp: #1]}}
    \newcommand{\js}[1]{{\textbf{\color{olive}[JS: #1]}}}
\else
    \newcommand{\jl}[1]{}
    \newcommand{\xlc}[1]{}
    \newcommand{\ba}[1]{}
    \newcommand{\jp}[1]{}
    \newcommand{\js}[1]{}
\fi

\ifshowcomments
    \title{\color{red}COMMENTS ARE DISPLAYED: Exploring the Uncertainty Properties of Neural Networks’ Implicit Priors in the Infinite-Width Limit}
\else
    \title{Exploring the Uncertainty Properties of Neural Networks’ Implicit Priors in the Infinite-Width Limit}
\fi

\author{%
  Ben Adlam\thanks{Authors contributed equally to this work. \textsuperscript{\textdagger}Work done as a member of the Google AI Residency program (https://g.co/airesidency).}~~\textsuperscript{\textdagger}
  \quad Jaehoon Lee~\textsuperscript{*}
  \quad Lechao Xiao~\textsuperscript{*}
  \quad Jeffrey Pennington
  \quad Jasper Snoek\\
  Google Brain\\
  \texttt{\{adlam, jaehlee, xlc, jpennin, jsnoek\}@google.com} \\
}

\iclrfinalcopy
\begin{document}

\maketitle

\begin{abstract}
Modern deep learning models have achieved great success in predictive accuracy for many data modalities. However, their application to many real-world tasks is restricted by poor uncertainty estimates, such as overconfidence on out-of-distribution (OOD) data and ungraceful failing under distributional shift. Previous benchmarks have found that ensembles of neural networks (NNs) are typically the best calibrated models on OOD data. Inspired by this, we leverage recent theoretical advances that characterize the function-space prior of an ensemble of infinitely-wide NNs as a Gaussian process, termed the neural network Gaussian process (NNGP). We use the NNGP with a softmax link function to build a probabilistic model for multi-class classification and marginalize over the latent Gaussian outputs to sample from the posterior. This gives us a better understanding of the implicit prior NNs place on function space and allows a direct comparison of the calibration of the NNGP and its finite-width analogue. We also examine the calibration of previous approaches to classification with the NNGP, which treat classification problems as regression to the one-hot labels. In this case the Bayesian posterior is exact, and we compare several heuristics to generate a categorical distribution over classes. We find these methods are well calibrated under distributional shift. Finally, we consider an infinite-width final layer in conjunction with a pre-trained embedding. This replicates the important practical use case of transfer learning and allows scaling to significantly larger datasets. As well as achieving competitive predictive accuracy, this approach is better calibrated than its finite width analogue.
\end{abstract}

\section{Introduction}
\jl{Mention NTK when appropriate and say that results are similar. Some comparison in the appendix.}

% \jp{Perhaps merge this and the next paragraph and get to the point a bit faster.}
% \js{Agreed, the ICLR community knows deep nets are prevalent}
% The arrival of large, neural network (NN) based models has generated substantial progress in many established machine learning tasks such as computer vision, speech recognition, and machine translation. When provided with many examples, these methods seem able to generate high quality representations of complex data with little need for sophisticated domain knowledge. 
% In fact, the current paradigm prescribes that the data should speak for themselves, in contrast to the hand-engineered features of the past. On standard supervised learning benchmarks, this approach has led to a step-change in accuracy and meant that new data modalities are accessible to machine learning.
%The high profile nature of these findings and their substantial media coverage has generated considerable interest in applying these techniques in new areas that were previously limited by complexity of the data or the insufficient accuracy of more classical methods.\js{This last sentence is probably unnecessary embellishment}
% However, while these models have demonstrated remarkable predictive performance on test data drawn from the same distribution as their training data, the demands of real world applications often require stringent levels of robustness to novel, changing, or shifted distributions.

% \js{Deep neural networks (NNs) have achieved state-of-the-art accuracy across many data modalities, but 

Large, neural network (NN) based models have demonstrated remarkable predictive performance on test data drawn from the same distribution as their training data, but the demands of real world applications often require stringent levels of robustness to novel, changing, or shifted distributions.
% The representations learned by NNs are difficult to interpret and able to fit spurious correlations. Therefore, if these models are ever to be widely applied in high-risk areas like medicine or autonomous driving, they must gain our trust in other ways. 
Specifically, we might ask that our models are calibrated.  Aggregated over many predictions, well calibrated models report confidences that are consistent with measured performance. \js{Not sure we need to mention metrics here.}\jl{removed}
The Brier score (BS), expected calibration error (ECE), and negative log-likelihood (NLL) are common measurements of calibration \citep{brier1950verification,naeini2015obtaining,gneiting2007strictly}. 

Empirically, there are many concerning findings about the calibration of deep learning techniques, particularly on \emph{out-of-distribution} (OOD) data whose distribution differs from that of the training data. For example,~\citet{mackay-1992a} showed that non-Bayesian NNs are overconfident away from the training data and \citet{hein-2019} confirmed this theoretically and empirically for deep NNs that use ReLU. For in-distribution data, {\em post-hoc} calibration techniques such as temperature scaling tuned on a validation set~\citep{platt1999probabilistic,guo-2017} often give excellent results; however, such methods have not been found to be robust on shifted data and indeed sometimes even reduce calibration on such data \citep{ovadia-19}. Thus finding ways to detect or build models that produce reliable probabilities when making predictions on OOD data is a key challenge.

Sometimes data can be only slightly OOD or can shift from the training distribution gradually over time. This is called \emph{dataset shift}~\citep{Quionero-Candela-2009} and is important in practice for models dealing with seasonality effects, for example. While perfect calibration under arbitrary distributional shift is impossible, simulating plausible kinds of distributional shift that may occur in practice at different intensities can be a useful tool for evaluating the calibration of existing models. A recently proposed benchmark takes this approach~\citep{ovadia-19}. For example, using several kinds of common image corruptions applied at various intensities, the authors observed the degradation in accuracy expected of models trained only on clean images~\citep{hendrycks2019benchmarking,mu2019mnist}, but also saw very different levels of calibration, with deep ensembles~\citep{Lakshminarayanan2017} proving the best.

\paragraph{Bridging Bayesian Learning and Neural Networks} In principle, Bayesian methods provide a promising way to tackle calibration, allowing us to define models with and infer under specific aleatory and epistemic uncertainty. Typically, the datasets on which deep learning has proven successful have high SNR, meaning epistemic uncertainty is dominant and model averaging is crucial because our overparameterized models are not determined by the training data. Indeed \citet{wilson2020case} argues that ensembles are a kind of Bayesian model average. 

\js{It might be worth stating that Bayesian methods give nice guarantees of generalization via PAC Bayes and calibration via the Bernstein von Mises Theorem. Also, the GP construction is the only one currently known that allows us to *exactly* compute the posterior of a Bayesian NN.}
%Moreover, many recent computational innovations have made Bayesian methods more practical \citep{}.

Ongoing theoretical work has built a bridge between NNs and Bayesian methods \citep{neal, lee2018deep, matthews2018}, by identifying NNs as converging to Gaussian processes in the limit of very large width. Specifically, the neural network Gaussian process (NNGP) describes the prior on function space that is realized by an i.i.d. prior over the parameters. The function space prior is a GP with a specific kernel that is defined recursively with respect to the layers. While the many heuristics used in training NNs may obfuscate the issue, little is known theoretically about the uncertainty properties implied by even basic architectures and initializations of NNs. Indeed theoretically understanding overparameterized NNs is a major open problem. With the NNGP prior in hand, it is possible to disambiguate between the uncertainty properties of the NN prior and those due to the specific optimization decisions by performing Bayesian inference. Moreover, it is only in this infinite-width limit that the posterior of a Bayesian NN can be computed exactly.

\subsection{Summary of contributions}
This work is the first extensive evaluation of the uncertainty properties of infinite-width NNs. Unlike previous work, we construct a valid probabilistic model for classification tasks using the NNGP, i.e.\ a label's prediction is always a categorical distribution. We perform neural network Gaussian process classification (NNGP-C) using a softmax link function to exactly mirror NNs used in practice. We perform a detailed comparison of NNGP-C against its corresponding NN on clean, OOD, and shifted test data and find NNGP-C to be significantly better calibrated and more accurate than the NN.\js{What is meant exactly by more performant?} \jl{We mean better accuracy?}

Next, we evaluate the calibration of neural network Gaussian process regression (NNGP-R) on both UCI regression problems and classification on CIFAR10. As the posterior of NNGP-R is a multivariate normal and so not a categorical distribution, a heuristic must be used to calculate confidences for classification problems. On the full benchmark of~\citet{ovadia-19}, we compare several such heuristics, and against the standard RBF kernel and ensemble methods. We find the calibration of NNGP-R to be competitive with the best results reported in~\citep{ovadia-19}. However in the process of preparing our findings for publication, newer strong baselines have been reported that we do not compare against.\footnote{See https://www.github.com/google/uncertainty-baselines.}

Finally, we consider NNs whose last layer only is infinite by taking a pre-trained embedding and using an infinite-width final layer (abbreviated as NNGP-LL). This mirrors an important practical use case for practitioners unable to retrain large models from scratch, but looking to adapt one to their particular data, which could have relatively few samples. We compare the calibration of NNGP-LL with a multi-layer FC network, fine-tuning of the whole network, and the gold standard ensemble method, and again we find NNGP-LL to have competitive calibration. While NNGP-LL is not a principled Bayesian method, it allows scaling to much larger datasets and potential synergy with recent advances in un- and semi-supervised learning due to the pre-training of the embedding, making this method applicable to many real-world uncertainty critical applications like medical imaging.

\section{Background}
\label{sec:background}
% \xlc{Embedding + NNGP-C: is this considered a fully - Bayesian model ? One upshot is much fewer training data is needed at least this is true for regression. }\jl{Hard to argue being fully Bayesian but maybe partial? It would be interesting to look at partial benchmark on partial dataset, 10k/5k/1k maybe? Lechao saw accuracy don't drop as much even if you use subset, but what about uncertainty measures..? }

\citet{nealthesis} identified the connection between infinite width NNs and Gaussian processes, showing that the outputs of a randomly initialized one-hidden layer NN converges to a Gaussian process as the number of units in the layer approaches infinity. 
% As NNs are often structured as affine transformations followed by pointwise applications of nonlinearities, 
Let $z_i^l(x)$ describe the $i$\textsuperscript{th} pre-activation following a linear transformation in the $l$\textsuperscript{th} layer of a NN. At initialization, the parameters of the NN are independent and random, so the central-limit theorem can be used to show that the pre-activations become Gaussian with zero mean and a covariance matrix $\K(x, x') = \mathbb E[z^l_i(x)z^l_i(x')]$.% 
% \xlc{Added}
\footnote{Note that $\K$ is independent of $i$, due to the independence between $z^l_{i}$ and $z^l_{i'}$ for $i\neq i'$. In other words, the covariances of the GP have a Kronecker factorization $\K\otimes {\bf \text I}_c$, where $c$ is the number of classes.}

Knowing the distributions of the outputs, one can apply Bayes theorem to compute the posterior distribution for new observations, which we detail in Sec.~\ref{sec:NN-GPC} for classification and Sec.~\ref{sec:NN-GPR} for regression. Moreover, when the parameters evolve under gradient flow to minimize a loss, the output distribution\footnote{The source of randomness comes from random initialization.} remains a GP but with a different kernel called the neural tangent kernel (NTK) \citep{Jacot2018ntk, lee2019wide}. This allows us to derive the exact formula for \emph{ensemble training} of infinite width networks; see Sec~\ref{sec:nngp and ntk description} for more details.
While we analyzed uncertainty properties of predictive output variance of NTK in conjunction, we observed that results and trends follow that of the NNGP.

These observations have recently been significantly generalized to include NNs with more than one hidden layer~\citep{lee2018deep, matthews2018} and with a variety of architectures, including weight-sharing (CNNs)~\citep{xiao18a, novak2018bayesian, garriga2018deep}, skip-connections, dropout \citep{schoenholz2016,pretorius2019expected}, batch norm \citep{yang2019mean}, weight-tying (RNNs)\citep{yang2019tensor}, self-attention \citep{hron2020}, graphical networks \citep{du2019graph}, and others. In this work, we focus on FC and CNN-Vec NNGPs, whose kernels are derived from fully-connected networks and convolution networks without pooling respectively (see Sec~\ref{sec:nngp and ntk description} for precise definitions). When it is required, we prepend FC- or C- to the NNGP to distinguish between these two variants. We use the Neural Tangents library of~\citet{novak2019neural} to automate the transformation of finite-width NNs to their corresponding Gaussian processes.

Another line of work has connected NNs and GPs by combining them in a single model. \citet{hinton2008using} trained a deep generative model and fed its output in to a GP for discriminative tasks. \citet{wilson2016a} consider a similar approach, creating a GP whose kernel is given as the composition of a nonlinear mapping (\emph{e.g.} a NN) and a spectral kernel. They scale this approach to larger datasets in \citep{wilson2016stochastic} using variational inference, but \citep{tran2018calibrating} found this method to be poorly calibrated.

\section{Full Bayesian Treatment of Classification with Neural Kernels}
\label{sec:NN-GPC}
It is common to interpret the logits of a NN once mapped through a softmax as a categorical distribution over the labels for each point. Indeed cross entropy loss is sometimes motivated as the KL divergence between the predicted distribution and the observed label. Similarly, while the initialization scheme used for a NN's parameters is often chosen for optimization reasons, it can also be thought of as a prior. This implicit prior over functions and over the distribution of labels has effects, despite the decision of most common training algorithms in deep learning to forgo explicitly trying to find its posterior. In this section, we take seriously this implicit prior and utilize the simple characterization it has over logits in the infinite width limit to define a probabilistic model for mutli-class classification as
\eq{\label{eq_gpc_def}
    y \sim \softmax(f(x)), \text{ where } f \sim \GP(\mathbf{0}, \K),
}
where $\K$ is the NNGP kernel. If the prior is a correct model for the data generation process, then the posterior is optimal for inference. Thus, by avoiding heuristic approaches to inference, we are able to directly evaluate the prior. Then by comparing this to more standard gradient-based training methods, we can understand their effect on calibration. 

For training data $(\X, \Y)$, the posterior on a test point $x$ can be found by marginalizing out the latent space. Denote $F\deq f(\X)$, $f\deq f(x)$, and $\mathbf{F}$ as the concatenation of $F$ and $f$, then
\eq{
    p(y|\X,\Y) = \int \softmax(f) p(f|\Y) \diff f,
}
where
\eq{
    p(\mathbf{F} | \Y) = p(\Y|F) p(\mathbf{F}) / p(\Y) \propto  \mathcal{N}(\mathbf{F};\mathbf{0}, \K(\mathbf{F}, \mathbf{F})) \prod_i \softmax(F_i)_{\Y_i}.
}
See \citep{williams2006gaussian} for details. We generate samples from the joint posterior distribution of $f$ and $F$ using elliptical slice sampling (ESS) \citep{murray2010elliptical}. Note that the latent space dimension in the datasets we consider is substantial, which makes inference with ESS computationally intensive, especially with hyperparameter tuning of the kernel. Therefore, we focus our attention on FC and CNN-Vec kernels.

We found the performance of NNGP-C to be sensitive to the kernel hyperparameters. To tune these parameters we used the Google Vizier service \citep{golovin2017google} with a budget of 250 trials and selected the setting with the best log-likelihood on a validation set. We use the same hyperparamters for the NN to make a direct comparison of the prior. The additional hyperparameters required for the NN, like width and training time, were also tuned using Vizier. We also compared against the NN performance when all its hyperparameters are tuned, and found the accuracy of the NN improved but the calibration results were broadly similar. See the supplement for additional details. 

\begin{figure}
    \centering
    \includegraphics[width=0.32\linewidth]{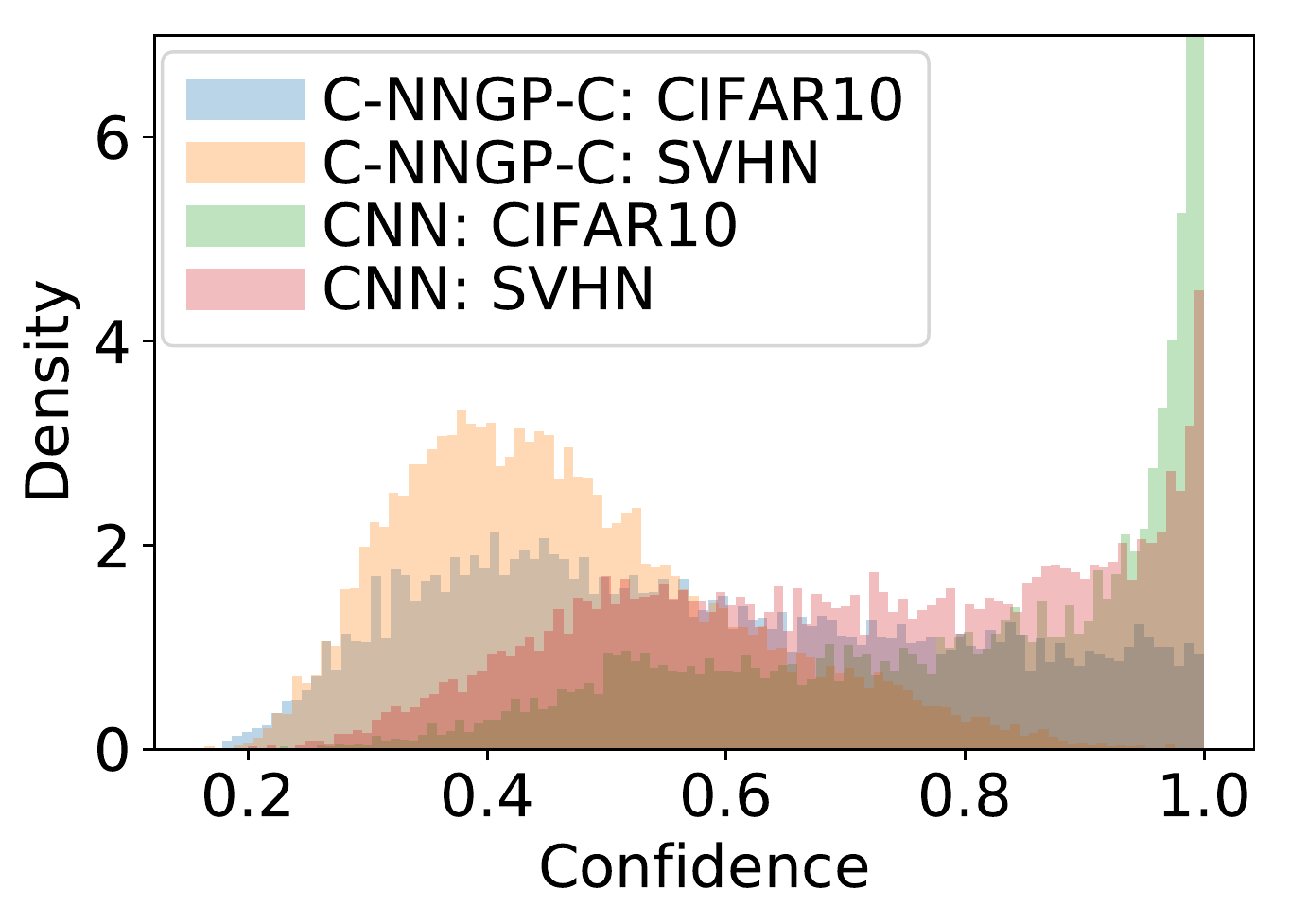}
    \includegraphics[width=0.32\linewidth]{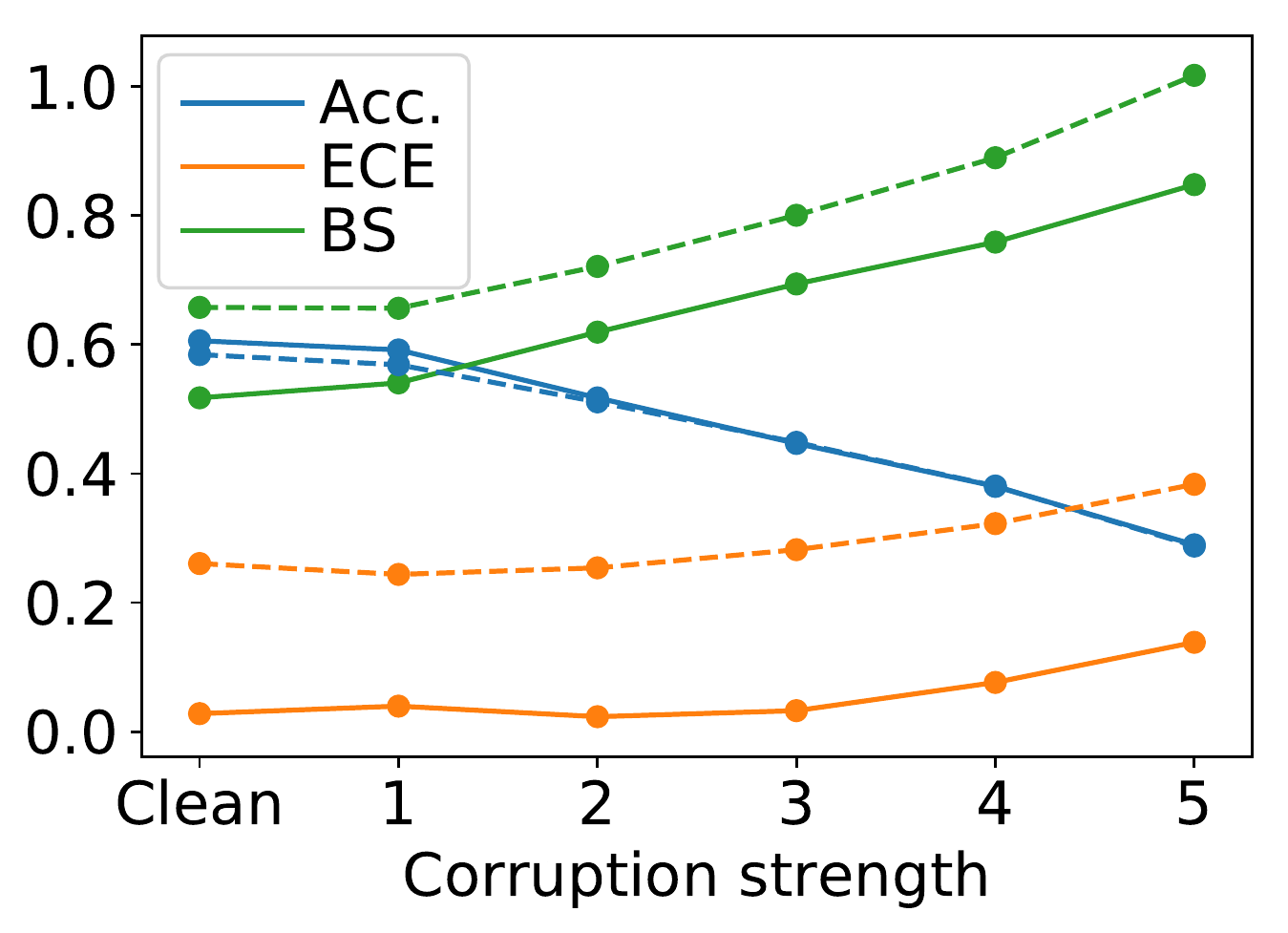}
    \includegraphics[width=0.32\linewidth]{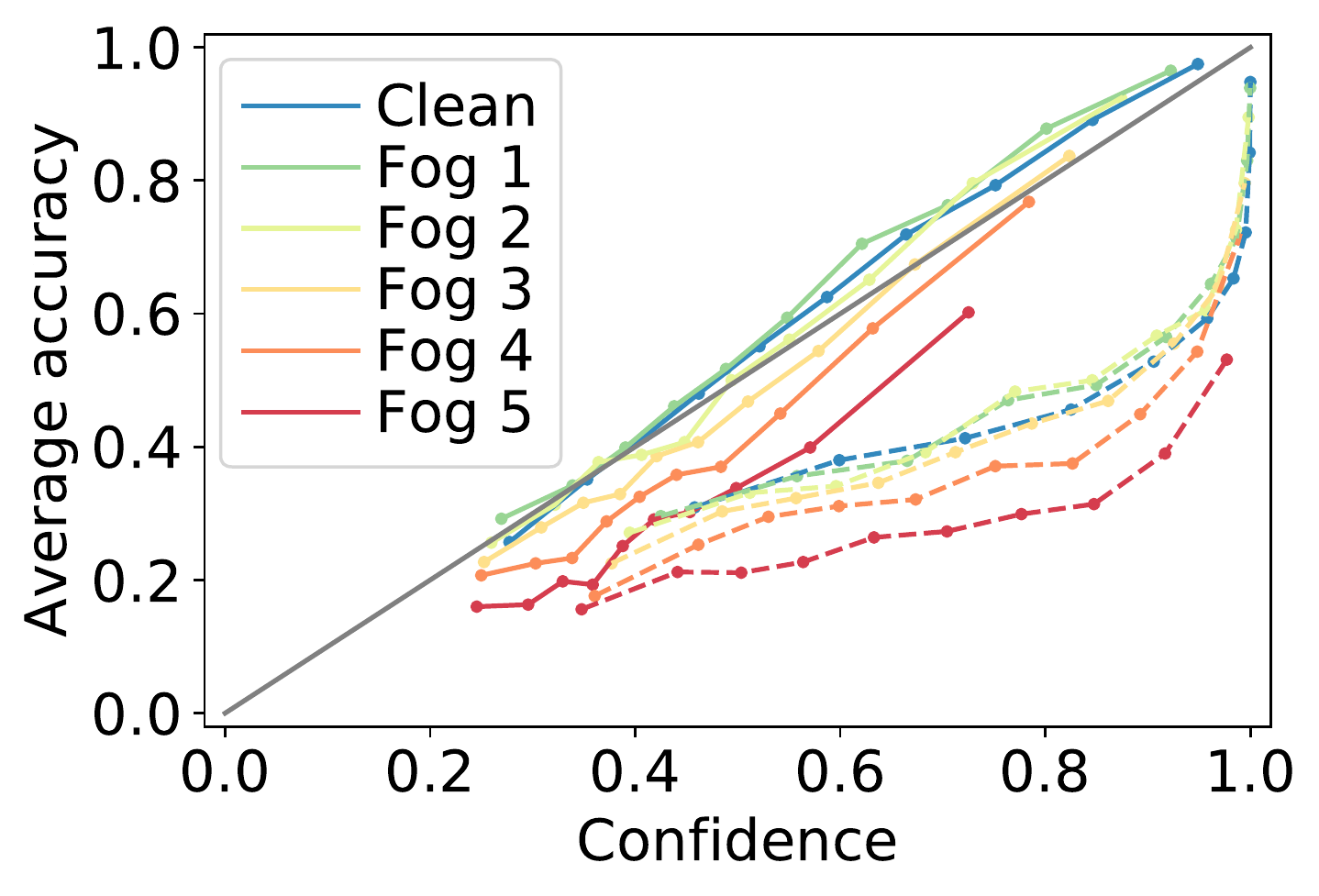}
    \caption{Investigating the calibration of Gaussian process classification with CNN-GP kernels. %Eq.~\ref{eq_gpc_def} for CNN-Vec.
    (\textbf{left}) Histogram of the confidence of the posterior distribution for each test point. We compare the C-NNGP-C and a finite width CNN on an identically distributed test set (CIFAR10) and an OOD test set (SVHN). C-NNGP-C shows lower confidence and higher entropy on both test sets compared to the CNN, which has very high confidence on many points as indicated by the spike around 1. On the clean data the CNN's overconfidence hurts its calibration, as it achieves worse ECE and BS than C-NNGP-C. (\textbf{middle}) The performance of both models, C-NNGP-C is solid and CNN dashed, under increasing distributional shift given by the CIFAR10 fog corruption. The accuracy of the CNN and C-NNGP-C are comparable as the shift intensity increases, but C-NNGP-C remains better calibrated throughout. (\textbf{right}) We bin the test set into ten bins sorted by confidence, and we plot mean confidence against mean accuracy. C-NNGP-C remains closer to $x=y$ than the CNN, which shifts toward increasing overconfidence.}
    \label{fig_GPC}
\end{figure}

Our main findings for NNGP-C, summarized in Fig.~\ref{fig_GPC} and Table~\ref{tab:gpc}, show that NNGP-C is well calibrated and outperforms the corresponding NN. This indicates that, rather than the prior induced by the random initialization scheme, the MAP-based training of the NN is partly responsible for its poor calibration.

% This indicates that the MAP-based training of the NN is partly responsible for its poor calibration and helps explain the success of ensembles.

\begin{table}[]
\centering
\caption{Comparing NNs and against the equivalent NNGP-C on CIFAR-10 and evaluated on several test sets.  We observe that the NNGP-C outperforms its parametric NN counterpart on \emph{every} metric. We see particularly significant improvements in ECE and NLL, implying that NNGP-C is considerably better calibrated. For complete results on other intensities of Fog corruption see Table~\ref{tab:sm_gpc}. \jl{Restructured the table so that we do not use the whole page. }\ba{Remove fog 2, 3, and 4. Put full table in SM.}}
\vspace{0.1cm}
\label{tab:gpc}
\renewcommand{\arraystretch}{0.8}% Tighter
\small{
\begin{tabular}{@{}ll|cc|cc@{}}
\toprule
Data      &  Metric   & FC-NN & FC-NNGP-C & CNN   & C-NNGP-C \\ \midrule\midrule
\multirow{4}{*}{CIFAR10}  & ECE & 0.209 & \textbf{0.072}   & 0.283 & \textbf{0.031}  \\
                          & Brier Score  & 0.711 & \textbf{0.629}   & 0.685 & \textbf{0.519}  \\
                          & Accuracy  & 0.487 & \textbf{0.518}   & 0.576 & \textbf{0.609}  \\
                          & NLL & 17383 & \textbf{14007}   & 21786 & \textbf{11215}  \\
                          \midrule
\multirow{4}{*}{Fog 1}    & ECE & 0.178 & \textbf{0.098}   & 0.271 & \textbf{0.042}  \\
                          & Brier Score  & 0.707 & \textbf{0.655}   & 0.685 & \textbf{0.542}  \\
                          & Accuracy  & 0.471 & \textbf{0.496}   & 0.561 & \textbf{0.590}  \\
                          & NLL & 16702 & \textbf{14671}   & 19716 & \textbf{11755}  \\
                          \midrule
% \multirow{4}{*}{Fog 2}    & ECE & 0.177 & \textbf{0.073}   & 0.286 & \textbf{0.018}  \\
%                           & Brier Score  & 0.758 & \textbf{0.711}   & 0.747 & \textbf{0.620}  \\
%                           & Accuracy  & 0.416 & \textbf{0.425}   & 0.507 & \textbf{0.519}  \\
%                           & NLL & 18026 & \textbf{16233}   & 20822 & \textbf{13819}  \\
%                           \midrule
% \multirow{4}{*}{Fog 3}    & ECE & 0.202 & \textbf{0.039}   & 0.320 & \textbf{0.03}  \\
%                           & Brier Score  & 0.822 & \textbf{0.759}   & 0.836 & \textbf{0.694}  \\
%                           & Accuracy  & 0.358 & \textbf{0.363}   & 0.445 & \textbf{0.445}  \\
%                           & NLL & 20092 & \textbf{17638}   & 23752 & \textbf{16019}  \\
%                           \midrule
% \multirow{4}{*}{Fog 4}    & ECE & 0.236 & \textbf{0.026}   & 0.352 & \textbf{0.071}  \\
%                           & Brier Score  & 0.881 & \textbf{0.797}   & 0.914 & \textbf{0.758}  \\
%                           & Accuracy  & 0.307 & \textbf{0.311}   & 0.385 & \textbf{0.380}  \\
%                           & NLL & 22589 & \textbf{18867}   & 27055 & \textbf{18353}  \\
%                           \midrule
\multirow{4}{*}{Fog 5}    & ECE & 0.279 & \textbf{0.057}   & 0.420 & \textbf{0.134}  \\
                          & Brier Score  & 0.961 & \textbf{0.847}   & 1.052 & \textbf{0.846}  \\
                          & Accuracy  & 0.241 & \textbf{0.251}   & 0.287 & \textbf{0.289}  \\
                          & NLL & 26345 & \textbf{20694}   & 33891 & \textbf{22014}  \\
                          \midrule
\multirow{2}{*}{SVHN}     & Mean Confidence & 0.537 & 0.335   & 0.718 & 0.463  \\
                          & Entropy & 1.230 & 1.840   & 0.733 & 1.403  \\
                          \midrule
\multirow{2}{*}{CIFAR100} & Mean Confidence & 0.651 & 0.398   & 0.812 & 0.474  \\
                          & Entropy & 0.944 & 1.663   & 0.493 & 1.420  \\ 
\end{tabular}
}

\end{table}

\section{Regression with the NNGP}
\label{sec:NN-GPR}
As observed in Sec.~\ref{sec:NN-GPC}, inference with NNGP-C is challenging as the posterior is intractable. In this section, we consider Gaussian process regression using the NNGP (abbreviated as NNGP-R), which is defined by the model
\eq{\label{eq_gpr_def}
    y \sim f(x) + \e, \text{ where } f \sim \GP(\mathbf{0}, \K),
}
where $\K$ is the NNGP kernel and $\e\sim \mathcal N(0, \sigma_{\epsilon}^2)$ is an independent noise term. One major advantage of NNGP-R is that the posterior is analytically tractable. The posterior at a test point $x$ has a Gaussian distribution with mean and variance given by
\eq{\label{eq_gpr_mean_var}
    \mu(x) = \K(x, \X)\K_{\epsilon}(\X, \X)^{-1}\mathcal{Y} \quad \text{and} \quad 
    \sigma^2(x) = \K(x, x) - \K(x, \X) \K_{\epsilon}(\X, \X)^{-1} \K(\X, x)
}
where $(\X, \Y)$ is the training set of inputs and targets respectively and $\K_{\epsilon} \equiv \K +\sigma_\epsilon^2 { \text{\bf I}} $.
Since $\K$ has a Kronecker factorization, the complexity of inference is $\mathcal O(|\X|^3)$ rather than $\mathcal O(c^3|\X|^3)$.
For regression problems where $y\in \R^d$, the variance describes the model's uncertainty about the test point. Note that while this avoids the difficulties of approximate inference methods like MCMC, computation of the kernel inverse means running time scales cubically with the dataset size.

% Recall that for an infinite-width neural network the outputs is a GP
% \eq{\label{eq_gpr_def}
%     y \sim f(x) + \e, \text{ where } f \sim \GP(\mathbf{0}, \K),
% }
% where $\K$ is the NNGP kernel and $\e\sim \mathcal N(0, \sigma_{\epsilon})$ is an independent noise term. The posterior at a test point $x$ is exactly tractable and has a Gaussian distribution with mean and variance 
% \eq{\label{eq_gpr_mean_var}
%     \mu(x) = \K(x, \X)\K_{\epsilon}(\X, \X)^{-1}\mathcal{Y} \quad \text{and} \quad 
%     \sigma^2(x) = \K(x, x) - \K(x, \X) \K_{\epsilon}(\X, \X)^{-1} \K(\X, x)
% }
% where $(\X, \Y)$ is the training set of inputs and targets respectively and $\K_{\epsilon} \equiv \K +\sigma_\epsilon^2 { \text{\bf I}} $.
% For regression problems where $y\in \R^K$, this posterior variance term can readily be interpreted as the model's uncertainty about the test point and the negative log likely (NLL) can be computed exactly. 

\subsection{Benchmark on UCI Datasets}

We perform non-linear regression experiments proposed by~\citet{hernandez2015probabilistic}, which is a standard benchmark for evaluating uncertainty of Bayesian NNs. We use all datasets except for \texttt{Protein} and \texttt{Year}. Each dataset is split into 20 train/test folds. We report our result in Table~\ref{tab:uci}, comparing against the following strong baselines: Probabilistic BackPropagation with the Matrix Variate Gaussian distribution (PBP-MV)~\citep{sun2017learning}, Monte-Carlo Dropout~\citep{gal2016dropout} evaluated with hyperparameter tuning as done in~\citet{mukhoti2018importance} and Deep Ensembles~\citep{Lakshminarayanan2017}. We also evaluated a standard GP with RBF kernel $K_\text{RBF} (x, x')= \beta \exp\left(-\gamma ||x - x'||^2 \right)$ for comparison.

Instead of maximizing train NLL for model selection, we performed hyperparameter search on a validation set (we further split the training set so that overall train/valid/test split is 80/10/10), as commonly done in NN model selection and in the BNN context applied in~\citep{mukhoti2018importance}. For NNGP-R, the following hyperparameters were considered.  The activation function is chosen from (ReLU, Erf), number of hidden layers among [1, 4, 16],  $\sigma_w^2$ from [1, 2, 4], $\sigma_b^2$ from [0., 0.09, 1.0], readout layer weight and bias variance are chosen either same as body or $(\sigma_w^2, \sigma_b^2) = (1, 0)$. For the GP with RBF kernel we evaluated $\gamma \in \{10^{k}: k \in [-5, -4, ... , 3]\}$ and $\beta  \in  \{10^{k}: k \in [-3, -4, ... , 3]\}$. For both experiments, $\sigma_\e\in$ \texttt{np.logspace(-6, 4, 20)}.

We found that NNGP-R can outperform and remain competitive with existing methods in terms of both root-mean-squared-error (RMSE) and negative-log-likelihood (NLL). In Table~\ref{tab:uci}, we observe that NNGP-R achieves the lowest RMSE on the majority (5/8) of the datasets and competitive NLL. When using the NTK as the GP's kernel instead, we saw broadly similar results.

% \jl{Do we want to say more?}\js{+1 e.g. From Table~\ref{tab:uci}, we observe that the NNGP achieves the lowest RMSE on the majority (5/8) of the datasets and competitive NLL.}

\begin{table}[ht!]
\centering
\caption{Result for regression benchmark on UCI Datasets. Note $\pm x$ reports the standard error around estimated mean for 20 splits. We compare to strong baselines: PMP-MV~\citep{sun2017learning}, MC-Dropout from \citet{mukhoti2018importance} and Deep Ensembles~\citep{Lakshminarayanan2017}}
\label{tab:uci}
\caption*{Average RMSE Test Performance}
\resizebox{\textwidth}{!}{%
\renewcommand{\arraystretch}{1.0}% Tighter
\begin{tabular}{@{}lcllllll@{}}
\toprule
Dataset             & $(m, d)$    & PBP-MV %\citep{sun2017learning} 
& Dropout %\citep{mukhoti2018importance}  
&    Ensembles %\citep{Lakshminarayanan2017} 
& RBF           &  FC-NNGP-R        \\ \midrule\midrule
Boston Housing      &(506, 13)   & 3.11 $\pm$ 0.15 & \textbf{2.90 $\pm$ 0.18} & 3.28 $\pm$ 1.00    & 3.24  $\pm$ 0.21 & 3.07 $\pm$ 0.24 \\
Concrete Strength   &(1030, 8)   & 5.08 $\pm$ 0.14 & \textbf{4.82 $\pm$ 0.16} & 6.03 $\pm$ 0.58    & 5.63 $\pm$ 0.24  &  5.25 $\pm$ 0.20 \\
Energy Efficiency   &(768, 8)    & \textbf{0.45 $\pm$ 0.01} & 0.54 $\pm$ 0.06 & 2.09 $\pm$ 0.29    & 0.50 $\pm$ 0.01  & 0.57 $\pm$ 0.02 \\
Kin8nm              &(8192, 8)   & \textbf{0.07 $\pm$ 0.00} & 0.08 $\pm$ 0.00 & 0.09 $\pm$ 0.00    & \textbf{0.07 $\pm$ 0.00}  & \textbf{0.07 $\pm$ 0.00} \\
Naval Propulsion    &(11934, 16) & \textbf{0.00 $\pm$ 0.00} &\textbf{0.00 $\pm$ 0.00} & \textbf{0.00 $\pm$ 0.00}    & \textbf{0.00 $\pm$ 0.00} & \textbf{0.00 $\pm$ 0.00} \\
Power Plant         &(9568, 4)   & 3.91 $\pm$ 0.04 & 4.01 $\pm$ 0.04 & 4.11 $\pm$ 0.17    & 3.82  $\pm$ 0.04  & \textbf{3.61 $\pm$ 0.04} \\
Wine Quality Red    &(1588, 11)  & 0.64 $\pm$ 0.01 & 0.62 $\pm$ 0.01 & 0.64 $\pm$ 0.04    & 0.64 $\pm$ 0.01 & \textbf{0.57 $\pm$ 0.01} \\
Yacht Hydrodynamics &(308, 6)    & 0.81 $\pm$ 0.06 & 0.67 $\pm$ 0.05 & 1.58 $\pm$ 0.48    & 0.60 $\pm$ 0.07 & \textbf{0.41 $\pm$ 0.04} \\ \bottomrule
\end{tabular}%
}
\vspace{0.3cm}
\caption*{Average Negative Log-Likelihood Test Performance}
\resizebox{\textwidth}{!}{%
\begin{tabular}{@{}lcllllll@{}}
\toprule
Boston Housing    &(506, 13)  &  \phantom{-}2.54 $\pm$  0.08      & {\bf \phantom{-}2.40 $\pm$ 0.04} & \phantom{-}2.41 $\pm$ 0.25 & \phantom{-}2.63 $\pm$ 0.09          & \phantom{-}2.65 $\pm$ 0.13     \\
Concrete Strength &(1030, 8)  & \phantom{-}3.04 $\pm$ 0.03      & \textbf{\phantom{-}2.93 $\pm$ 0.02} & \phantom{-}3.06 $\pm$ 0.18 & \phantom{-}3.52 $\pm$ 0.11       & \phantom{-}3.19 $\pm$ 0.05     \\
Energy Efficiency &(768, 8)   & \phantom{-}1.01 $\pm$ 0.01  & \phantom{-}1.21 $\pm$ 0.01     & \phantom{-}1.38 $\pm$ 0.22     & \textbf{\phantom{-}0.78 $\pm$ 0.06}     & \phantom{-}1.01 $\pm$ 0.04 \\
Kin8nm            &(8192, 8)  & \textbf{-1.28 $\pm$ 0.01} & -1.14 $\pm$ 0.01    & -1.20 $\pm$ 0.02    & -1.11 $\pm$ 0.01        & -1.15 $\pm$ 0.01    \\
Naval Propulsion   & (11934, 16)& -4.85 $\pm$ 0.06 & -4.45 $\pm$ 0.00 & -5.63 $\pm$ 0.05    & \textbf{-10.07 $\pm$ 0.01}  & -10.01 $\pm$ 0.01 \\
Power Plant       &(9568, 4)  & \phantom{-}2.78 $\pm$ 0.01  & \phantom{-}2.80 $\pm$ 0.01     & \phantom{-}2.79 $\pm$ 0.04 & \phantom{-}2.94 $\pm$ 0.01     & \textbf{\phantom{-}2.77 $\pm$ 0.02} \\
Wine Quality Red  &(1588, 11) & \phantom{-}0.97 $\pm$ 0.01      & \phantom{-}0.93 $\pm$ 0.01     & \phantom{-}0.94 $\pm$ 0.12     & -0.78 $\pm$ 0.07    & \textbf{-0.98 $\pm$ 0.06}    \\
Yacht Hydrodynamics &(308, 6)    & \phantom{-}1.64 $\pm$ 0.02  & \phantom{-}1.25 $\pm$ 0.01  & \phantom{-}1.18 $\pm$ 0.21 & \textbf{\phantom{-}0.49 $\pm$ 0.06}    & \phantom{-}1.07 $\pm$ 0.27   \\
\bottomrule
\end{tabular}%
}
\end{table}

\subsection{Classification as Regression}

Formulating classification as regression often leads to good results, despite being less principled~\citep{rifkin2003regularized, rifkin2004}. By doing so, we can compare exact inference for GPs to trained NNs on well-studied image classification tasks. Recently, various studies of infinite NNs have considered classification as regression tasks, treating the one-hot labels as independent regression targets (e.g.~\citep{lee2018deep, novak2018bayesian, garriga2018deep}). Predictions are then obtained as the argmax of the mean in Eq.~\ref{eq_gpr_mean_var}, \emph{i.e.} $\argmax_k \mu(x)_k$.\footnote{ While training NNs with MSE loss is more challenging, peak performance can be competitive with cross-entropy loss~\citep{lewkowycz2020large, hui2020evaluation,lee2020finite}.}

However, this approach does not provide confidences corresponding to the predictions. Note that the posterior gives support to all of $\R^d$, including points that are known to be impossible. Thus, a heuristic is required to extract meaningful uncertainty estimates from the posterior Eq.~\ref{eq_gpr_mean_var}, even though these confidences will not correspond to the Bayesian posterior of any model.

% To treat Bayesian probabilistic model of classification as regression problem, assume the Gaussian process regression (NNGP-R) is defined as 
% \eq{\label{eq_gpr_def}
%     y \sim f(x) + \e, \text{ where } f \sim \GP(\mathbf{0}, \K),
% }
% where $\K$ is the NNGP kernel and $\e\sim \mathcal N(0, \sigma_{\epsilon})$ is an independent noise term. The posterior at a test point $x$ is exactly tractable and has a Gaussian distribution with mean and variance 
% \eq{\label{eq_gpr_mean_var}
%     \mu(x) = \K(x, \X)\K_{\epsilon}(\X, \X)^{-1}\mathcal{Y} \quad \text{and} \quad 
%     \sigma^2(x) = \K(x, x) - \K(x, \X) \K_{\epsilon}(\X, \X)^{-1} \K(\X, x)
% }
% where $(\X, \Y)$ is the training set of inputs and targets respectively and $\K_{\epsilon} \equiv \K +\sigma_\epsilon^2 { \text{\bf I}} $. For regression problems where $y\in \R^d$, this posterior the variance term can readily be interpreted as the model's uncertainty about the test point.

% \subsection{Discussion on temperature scale}

% - Tempering as modifying prior.\\
% - RBF example\\
% - Generalize to NN kernels
% One of the earliest modern success in deep learning methods comes from image classification problems. Naturally lot of uncertainty bench-marking has been done on this context. 

Following \citep{albert1993bayesian, girolami2006variational}, we produce a categorical distribution for each test point $x$, denoted $p_x$, by defining 
\al{
    p_x(i)\deq \P[z_i = \max\h{y_1,\ldots, y_d}] = 
    \int  {\mathbf 1}  (i =\text{argmax}_j y_j)  \prod_{k=1}^Kp(y_k|x, \X, \Y) dy, \label{eq:appr-posterior}
}
where $(y_1,\ldots, y_d)$ is sampled from the posterior for $(x,y)$ and we used the independence of the posterior for each class.\footnote{This follows directly from the Kronecker structure of the NNGP and our treating the labels as independent regression targets.} Note that we also treat the predictions on different test points independently. In general, Eq.~\ref{eq:appr-posterior} does not have an analytic expression, and we resort to Monte-Carlo estimation. We refer readers to the supplement for comparison to other heuristics (e.g. passing the mean predictor through a softmax function and pairwise comparison). While this is heuristic, we find it is well calibrated (see Fig.~\ref{fig:gpr_box}). This is perhaps because the posterior Eq.~\ref{eq_gpr_mean_var} still represents substantial model averaging, and most uncertainty in high SNR cases is epistemic rather than aleatory.

\subsection{Benchmark on CIFAR10}

\begin{figure}[h]
    \centering
    \includegraphics[width=\columnwidth]{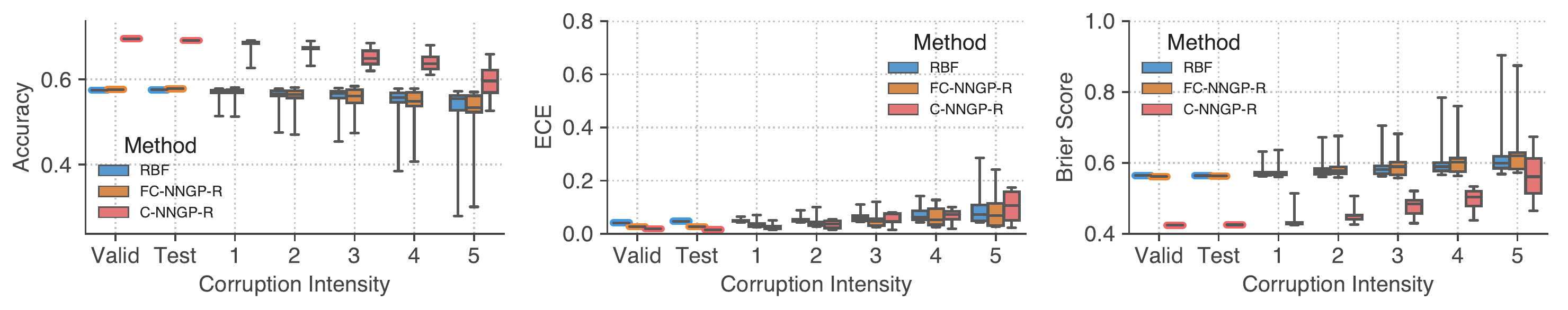}    
    \caption{Uncertainty metrics across shift levels on CIFAR10 using NNGP-R. CNN kernels perform best as well as being more robust to corruption. See the supplement for numerical values at each quartile as well as a comparison to the NTK (Fig.~\ref{fig:gpr_ntk_box}). All methods remain well calibrated for all intensities of shifts, with C-NNGP-R performing best, and significantly better than methods in \citep{ovadia-19}; contrast against their Fig.~2 and S4.}
    \label{fig:gpr_box}
\end{figure}

We examine the calibration of NNGP-R on increasingly corrupted images of CIFAR10-C \citep{hendrycks2019benchmarking} using the benchmark of \citep{ovadia-19}. The results are displayed in Fig.~\ref{fig:gpr_box}. While FC-NNGP is similar to the standard RBF kernel, the C-NNGP outperforms both in terms of calibration and accuracy. Moreover, we find that at severe corruption levels, the C-NNGP actually outperforms all methods in \citep{ovadia-19} (compare against their Table~G1) in BS and ECE.

\section{Bayesian or Infinite-Width Last Layer}
\label{sec:gpll}
As we have seen NNGP-C and NNGP-R are remarkably well-calibrated. However, obtaining high performing models can be computationally intensive, especially for large datasets. NNGP-C and NNGP-R have running times that are cubic in the dataset size, due to computation of the kernel's Cholesky decomposition, with NNGP-C suffering additionally from potentially slow convergence of MCMC. Moreover, performant NNGP kernels require substantial compute to obtain \citep{novak2018bayesian, arora2019on, novak2019neural, li2019enhanced} in contrast to training a NN to similar accuracies. Moreover, even though the most performant NNGP kernels are SotA for a kernel method \citep{Shankar2020NeuralKW}, they still under-perform SotA NNs by a large margin. 

To combine the benefits of the NNGP and NNs, obtaining models that are both performant and well calibrated, we propose stacking an infinite-width sub-network on top of a pre-trained NN. More precisely, we use features obtained from a pre-trained model as inputs for the NNGP. As such, the outputs of the combined network are drawn from a GP and we may use Eqs.~\ref{eq_gpr_mean_var} and \ref{eq:appr-posterior} for inference. We refer to this as neural network Gaussian process last-layer (NNGP-LL). Mathematically, the model is
\eq{\label{eq_gpll_def}
    y \sim f(g(x)) + \e, \text{ where } f \sim \GP(\mathbf{0}, \K),
}
where $g$ is a pre-trained embedding, $\K$ is the NNGP kernel, and $\e\sim \mathcal N(0, \sigma_{\epsilon})$ is a noise term. This draws inspiration from~\citep{hinton2008using, wilson2016a}, but with a \emph{multi-layer} NNGP kernel. Note we are specifically interested in the calibration properties and so the innovations in computational efficiency in \citep{wilson2016a} are complementary to our work.

This setup mirrors an important use case in practice, e.g. for customers of cloud ML services, who may use embeddings trained on vast amounts of non-domain-specific data, and then fine-tune this model to their specific use case. This fine tuning consists of either fitting a logisitic regression layer or deeper NNs using the embeddings obtained from their data, or perhaps training the whole NN generating the embedding by simply initializing with the pre-trained weights~\citep{kornblith2019better}. These strategies allow practitioners to obtain highly accurate models without substantial data or computation. However, little is understood about the calibration of these transfer learning approaches.

\begin{figure}[t!]
    \centering
    \includegraphics[width=\columnwidth]{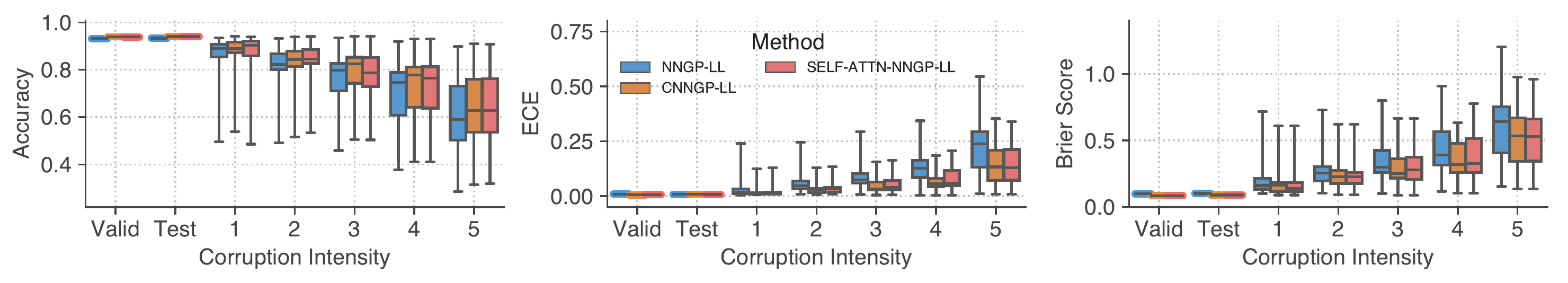}    
    \caption{Uncertainty metrics across shift levels on CIFAR10 using NNGP-LL with EfficientNet-B3 as an embedding. We observe that more complex neural kernels (convolution and self-attention NNGPs) can have higher performance while being more robust to corruptions. See Fig.~\ref{fig:gpll_embedding_ntk_box} for the comparison to the NTK.}
    \label{fig:gpll_embedding_box}
\end{figure}

We consider the EfficientNet-B3 \citep{tan2019efficientnet} embedding from TF-Hub\footnote{\texttt{https://www.tensorflow.org/hub}} and TF Keras Applictions\footnote{\texttt{https://www.tensorflow.org/api\_docs/python/tf/keras/applications}}  that is trained on ImageNet \citep{Deng2009Imagenet}, and perform our evaluations on CIFAR10 and its corruptions. 
For our experiments, we mainly use a multi-layer FC-NNGP as the top sub-network since its kernel is very fast to compute and the final FC layer of EfficientNet-B3 removes any spacial structure that might be exploited by convolutions. However, we also explore other kernel types (CNNs with pooling and self-attention layers) in Fig.~\ref{fig:gpll_embedding_box} by using earlier layers of EfficientNet-B3.
We compare our method with other popular last-layer methods for generating uncertainty estimates (Vanilla logisitic regression, using a deep NN for the last layers, temperature scaling \citet{platt1999probabilistic, guo-2017}, MC-Dropout \citep{gal2016dropout}, ensembles of several last-layer deep NNs). In the supplement we further investigate these results with a WideResNet \citep{zagoruyko2016wide} that we can train from scratch, using the initialization method in \citep{Douphin2019Meta}, which achieves good test performance on CIFAR-10 without BatchNorm \citep{ioffe2015batch}. This allows us to compare against the gold standard ensemble method. 

\begin{table}[ht!]
\centering
\caption{NNGP-LL with EfficientNet-B3 fine tuned on CIFAR10 as an embedding and evaluated over all corruptions and intensities. We show the quartiles for these evaluations for several different last-layer methods of obtaining uncertainties. Ensembles refers to~\citep{Lakshminarayanan2017} and Ens/Drp/T refers to combining Ensembles, MC-Dropout~\citep{gal2016dropout} and temperature scaling~\citep{guo-2017}. More specifically, we train a three-layer FC network with dropout using input from the embedding from EfficientNet-B3 and report results in the middle columns. The rightmost columns show that NNGP-LL can be well-calibrated with very little training data. See the supplement Fig.~\ref{fig:gpll_datasize} for a fine-grained box plots for each corruption level.}
\vspace{0.1cm}
\label{tab:gpll_datasize}
\resizebox{\textwidth}{!}{%
\begin{tabular}{l|ccc|ccccc}
\toprule
Method &     Vanilla & Ensembles & Ens/Drp/T & 100 & 1K & 5K & 10K & NNGP-LL \\
\midrule
Brier Score (25th)  &  0.230 &  0.218 &        0.182 &    0.363 &   0.256 &   0.218 &    0.203 &     \textbf{0.173} \\
Brier Score (50th)  &  0.351 &  0.331 &        \textbf{0.265} &    0.448 &   0.346 &   0.308 &    0.288 &     0.271 \\
Brier Score (75th)  &  0.521 &  0.511 &        0.410 &    0.572 &   0.478 &   0.436 &    0.409 &     \textbf{0.397} \\
\midrule
NLL (25th)          &  0.913 &  0.823 &        0.382 &    0.797 &   0.562 &   0.474 &    0.455 &     \textbf{0.367} \\
NLL (50th)          &  1.517 &  1.411 &        \textbf{0.569} &    0.991 &   0.746 &   0.682 &    0.655 &     0.586 \\
NLL (75th)          &  2.662 &  2.492 &        0.932 &    1.326 &   1.075 &   0.972 &    0.921 &     \textbf{0.905} \\
\midrule
ECE (25th)          &  0.104 &  0.098 &        \textbf{0.016} &    0.023 &   0.044 &   0.018 &    0.019 &     0.017 \\
ECE (50th)          &  0.160 &  0.154 &        0.028 &    0.040 &   0.062 &   \textbf{0.023} &    0.027 &     0.025 \\
ECE (75th)          &  0.247 &  0.243 &        0.079 &    0.081 &   0.104 &   \textbf{0.042} &    0.057 &     0.044 \\
\midrule
Accuracy (75th)     &  0.869 &  0.875 &        0.875 &    0.742 &   0.825 &   0.851 &    0.860 &     \textbf{0.884} \\
Accuracy (50th)     &  0.802 &  0.812 &        \textbf{0.813} &    0.674 &   0.758 &   0.784 &    0.798 &     \textbf{0.813} \\
Accuracy (25th)     &  0.714 &  0.719 &        0.718 &    0.582 &   0.663 &   0.689 &    0.704 &     \textbf{0.722} \\
\end{tabular}%
}
\end{table}

We find that in the transfer learning case, ensembles are quite ineffective alone. The best previous method is given by combining MC-Dropout, temperature scaling, and ensembles. However, this is still bested by NNGP-LL (see Table~\ref{tab:gpll_datasize}). We also examine the effect of the fine-tuning dataset size on calibration and performance. Remarkably, NNGP-LL is able to achieve accuracy and calibration comparable to ensembles with as few as 1000 training points. 
\xlc{accuracy is not quite comparable}

\section{Discussion}
% \begin{enumerate}
%     \item NN-GPC Calibrated, hard to scale up. Next step: investigate SOTA kernel?  
%     \item NN-GPR good for MSE task ?
%     \item Simple, efficient and practical way to obtain good calibration + accuracy; in particular for transfer learning ? Need a more thorough investigation.  
% \end{enumerate}

% \ba{TODO}

In this work, we explored several methods that exploit neural networks' implicit priors over functions in order to generate uncertainty estimates, using the corresponding Neural Network Gaussian Process (NNGP) as a means to harness the power of an infinite ensemble of NNs in the infinite-width limit. Using the NNGP, we performed fully Bayesian classification (NNGP-C) and regression (NNGP-R) and also examined heuristics for generating confidence estimates when classifying via regression. Across the board, we found that the NNGP provides good uncertainty estimates and generally delivers well-calibrated models even on OOD data. We found that NNGP-R is competitive with SOTA methods on the UCI regression task and remained calibrated even for severe levels of corruption. Despite their good calibration properties, as pure kernel methods, NNGP-C and NNGP-R cannot always compete with modern NNs in terms of accuracy. Adding an NNGP to the last-layer of a pre-trained model (NNGP-LL), allowed us to simultaneously obtain high accuracy and improved calibration. Moreover, we found NNGP-LL to be a simple and efficient way to generate uncertainty estimates with potentially very little data, and that it outperforms all other last-layer methods for generating uncertainties we studied. Overall, we believe that the infinite-width limit provides a promising direction to improve and better understand uncertainty estimates for NNs.

\subsubsection*{Acknowledgments}
We thank Jascha Sohl-Dickstein for feedback along the project and Roman Novak and Sam S. Shoenholz for the help on Neural Tangents~\citep{novak2019neural} library and JAX~\citep{jaxrepo}. Also we would like to further thank Roman Novak for detailed feedback on a manuscript draft.

We acknowledge the Python community~\citep{van1995python} for developing the core set of tools that enabled this work, including NumPy~\citep{numpy}, SciPy~\citep{scipy}, Matplotlib~\citep{matplotlib}, Pandas~\citep{pandas}, Jupyter~\citep{jupyter}, JAX~\citep{jaxrepo}, Neural Tangents~\citep{neuraltangents2019}, Tensorflow datasets and Google Colaboratory.
\small

\bibliography{references}
\bibliographystyle{iclr2021_conference}

\appendix
\newpage
\normalsize
\onecolumn
\clearpage
\appendix

\begin{center}
\textbf{\large Supplementary Material}
\end{center}

\setcounter{equation}{0}
\setcounter{figure}{0}
\setcounter{table}{0}
\setcounter{page}{1}
\setcounter{section}{0}

\renewcommand{\theequation}{S\arabic{equation}}
\renewcommand{\thefigure}{S\arabic{figure}}
\renewcommand{\thetable}{S\arabic{table}}

\section{Detailed Description of the NNGP and NTK}\label{sec:nngp and ntk description}
%\xlc{Adopted from \cite{lee2019wide}}\jl{Add NTK?}\xlc{Done!}\jl{Thanks!!}
In this section, we describe the FC-NNGP and the C-NNGP. Most of the contents are adopted from \cite{lee2018deep, novak2018bayesian, lee2019wide}, which we refer readers to for more technical details. 
\paragraph{NNGP:}
Let $\D \subseteq \mathbb R^{n_0} \times \mathbb R^{K}$ denote the training set and $\X=\left\{x: (x,y)\in \D\right\}$ and $\Y=\left\{y: (x,y)\in \D\right\}$ denote the inputs and labels, respectively. Consider a fully-connected feed-forward network with $L$
hidden
layers with widths $n_{l}$, for $l = 1, ..., L$ and a readout layer with $n_{L+1} = K$.  For each $x\in\mathbb R^{n_0}$, we use $h^l(x), x^l(x)\in\mathbb R^{n_l}$ to represent the pre- and post-activation functions at layer $l$ with input $x$. The recurrence relation for a feed-forward network is defined as 
\begin{align}
\label{eq:recurrence}
\begin{cases}
    h^{l+1}&=x^l W^{l+1} + b^{l+1}
    \\
    x^{l+1}&=\phi\left(h^{l+1}\right) 
    \end{cases}
    \,\, \textrm{and} 
    \,\,
    \begin{cases}
  W^{l}_{i, j}& = \frac {\sigma_\omega} {\sqrt{n_l}}  \omega_{ij}^l 
    \\
    b_j^l &= \sigma_b  \beta_j^l
\end{cases}
,
\end{align}
where $\phi$ is a point-wise activation function, $W^{l+1}\in \mathbb R^{n_l\times n_{l+1}}$ and $b^{l+1}\in\mathbb R^{n_{l+1}}$ are the weights and biases, $\omega_{ij}^l$ and $ b_j^l $ are the trainable variables, drawn i.i.d. from a standard Gaussian $ \omega_{ij}^l,  \beta_{j}^l\sim \mathcal N(0, 1)$ at initialization, and $\sws$ and $\sbs$ are weight and bias variances.

As the width of the hidden layers approaches infinity, the Central Limit Theorem (CLT) implies that the outputs at initialization $\left\{f(x)\right\}_{x\in\X}$ converge to a multivariate Gaussian in distribution. Informally, this occurs because the pre-activations at each layer are a sum of Gaussian random variables (the weights and bias), and thus become a Gaussian random variable themselves. 
See 
\cite{poole2016exponential,schoenholz2016, lee2018deep, xiao18a, yang2017} for more details, and \cite{matthews2018b_arxiv, novak2018bayesian}
for a formal treatment. 

Therefore, randomly initialized neural networks are in correspondence with a certain class of GPs (hereinafter referred to as NNGPs), which facilitates a fully Bayesian treatment of neural networks \citep{lee2018deep,matthews2018}. More precisely, let $f^{i}$ denote the $i$-th output dimension and $\infnngp$ denote the  sample-to-sample kernel function (of the pre-activation) of the outputs in the infinite width setting, 
\begin{align}
    \infnngp^{i, j}(x,x') = 
    \lim_{\min\pp{n_{1}, \dots, {n_L}}\to\infty}
    \mathbb E \left[ f^i(x)\cdot f^j(x')\right],
\end{align}
then $f(\X) \sim \mathcal{N}(0, \infnngp(\X, \X))$, where $\infnngp^{i, j}(x, x')$ 
denotes the covariance between the $i$-th output of $x$ and $j$-th output of $x'$, 
which can be computed recursively (see \citet[\S 2.3]{lee2018deep}.
For a test input $\tpoint\in \X_T$, the joint output distribution $f\left([\tpoint, \X]\right)$ is also multivariate Gaussian.
Conditioning on the training samples, $f(\X)=\Y$, the 
distribution of $\left.f(\tpoint)\right\vert \X, \Y$ is also a Gaussian $\mathcal N \left(\mu(\tpoint), \sigma^2(\tpoint)\right)$, 
\begin{align}
\label{eq:nngp-exact-posterior}
\mu(\tpoint) = \infnngp(\tpoint, \X) \infnngp^{-1}\Y, \quad
\sigma^2(\tpoint) = \infnngp(\tpoint, \tpoint) - \infnngp(\tpoint, \X) \infnngp^{-1}\infnngp(\tpoint, \X)^T,
\end{align}
and where $\infnngp = \infnngp(\X, \X)$.  
This is the posterior predictive distribution resulting from exact Bayesian inference in an infinitely-wide neural network.
\paragraph{C-NNGP:} 
The above arguments can be extended to convolutional architectures \cite{novak2018bayesian}.
By taking the number of channels in the hidden layers to infinity simultaneously, the outputs of CNNs also converge weakly to a Gaussian process (C-NNGP). The kernel of the C-NNGP takes into account the correlation between pixels in different spatial locations and can also be computed exactly via a recursively formula; \emph{e.g.}, see Eq.~(7) in \citep[\S 2.2]{novak2018bayesian}. Note that for convolutional architectures, there are two canonical ways of collapsing image-shaped data into logits. One is to vectorlize the image to a one-dimensional vector (CNN-Vec) and the other is to apply a global average pooling to the spatial dimensions (CNN-GAP). The kernels induced by these two approaches are very different and so are the C-NNGPs. We refer the readers to Section 3.2 of \citep[\S 3]{novak2018bayesian} for more details. In this paper, we have focused mostly on vectorization since it is more efficient to compute.
\paragraph{ATTN-NNGP:} There is also a correspondence between self-attention mechanisms and GPs. Indeed, for multi-head attention architectures, as the number of heads and the number of features tend to infinity, the outputs of an attention model also converge to a GP \citep{hron2020}. We refer the readers to \citet{hron2020} for technical details.    

\paragraph{NTK:} 
When neural networks are optimized using continuous gradient descent with learning rate $\eta$ on mean squared error (MSE) loss, the function evaluated on training points evolves as 
\begin{align}
    \partial_tf_t(\mathcal X) = -\eta J_t(\mathcal X)J_t(\mathcal X)^T \left(f_t(\mathcal X) - \mathcal{Y}\right)
\end{align}
where $J_t(\mathcal X)$ is the Jacobian of the output $f_t$ evaluated at $\mathcal X$ and $\Theta_t(\mathcal X,\mathcal X) = J_t(\mathcal X)J_t(\mathcal X)^T$ is the Neural Tangent Kernel (NTK). In the infinite-width limit, the NTK remains constant ($\Theta_t = \Theta$) throughout training \citep{Jacot2018ntk}. Thus the above equation is reduced to a constant coefficient ODE 
\begin{align}\label{eq: infinite-width-ode}
            \partial_tf_t(\mathcal X) = -\eta \infntk \left(f_t(\mathcal X) - \mathcal{Y}\right)
\end{align}
and the time-evolution of the outputs of unseen input $\X_T$ can be solved in closed form as a Gaussian with mean and covariance
% \footnote{Here {``\it+h.c.'' } is an abbreviation for ``plus the Hermitian conjugate''.}   
\begin{align}
      &\mu(\X_T) =\infntk\left(\X_T, \X\right)\infntk^{-1}\left(I -e^{- \eta \infntk t}\right)\Y \,,
      \label{eq:lin-exact-dynamics-mean}
      \\
      &\Sigma(\X_T, \X_T) =\infnngp\left(\X_T, \X_T\right) +\infntk(\X_T, \X)\infntk^{-1}\left(I-e^{- \eta \infntk t}\right) \infnngp \left(I - e^{-\eta \infntk t}\right) \infntk^{-1} \infntk\left(\X, \X_T \right)\nonumber \\
      &\phantom{\Sigma(\X_T, \X_T) =\infnngp\left(\X_T, \X_T\right)} -\left(\infntk(\X_T, \X)\infntk^{-1}\left(I-e^{- \eta \infntk t}\right) \infnngp\left(\X, \X_T \right) + h.c. \right).
    \label{eq:lin-exact-dynamics-var}
\end{align}
Note that the randomness of the solution is the consequence of the random initialization $f(\X) \sim \mathcal N(0, \K(\X, \X))$.  

Finally, the above arguments do not rely on the choice of architectures and we could likewisely define CNTK and ATTN-NTK, the NTK for CNNs and attention models, respectively.      
    
\clearpage
\section{Additional Figures for NNGP-C}

In this section, we show some additional plots and results comparing NNGP-C against standard NNs. Mainly, we address the method of hyperparameter tuning considered in the main text, where we fixed the hyperparameters that are common to both the NNGP and the NN, then only tuned the additional NN hyperparameters. Here, we show results for tuning all of the NN's hyperparameter from scratch.

\paragraph{Additional Tuning details.} 

For any tuning of hyperparameters, we split the original training set of CIFAR10 into a 45K training set and a 5K validation set. All models were trained using the 45K points, and we then selected the hyperparameters from the validation set performance. We introduced a constant that multiples the whole NNGP kernel, or equivalently scales the whole latent space vector or the last layer bias and weight standard deviations---we called this constant the kernel scale. For FC-NNGP-C, the activation function was tuned over $\{\relu, \erf\}$, the weight standard deviation was tuned over $[0.1, 2.0]$ on a linear scale, the bias standard deviation was tuned over $[0.1, 0.5]$ on a linear scale, the kernel scale was tuned over $[10^{-2}, 100]$ on a logarithmic scale, the depth was tuned over $\h{1,2,3,4,5}$, and the diagonal regularizer was tuned over $[0., 0.01]$ on a linear scale.

For the FC-NN, there are additional hyperparameters: the learning rate, training steps, and width. For the NN, we considered two types of tuning. Either, as in the main text, the hyperparameters that are shared with the NNGP are fixed and the additional hyperparameters are tuned, or, as we present in the supplement, all of the NN's hyperparameters are tuned from scratch. In either case, the activation, the weight standard deviation, the bias  standard deviation, the kernel scale, and the depth were tuned as above. The learning rate was tuned over $[10^{-4}, 0.1]$ on a logarithmic scale, the total training steps was tuned over $[10^{5}, 10^{7}]$ on a logarithmic, and the width was tuned over $\{64, 128, 256, 512\}$.

For the C-NNGP-C, all hypermaramters were treated as for the FC-NNGP-C case, except depth which was limited to $\h{1,2}$. For the CNN, we again considered the two types of tuning: either fixing common hyperparameters or retuning all hyperparameters from scratch. The CNN's learning rate was tuned over $[10^{-4}, 0.1]$ on a logarithmic scale, the total training steps was tuned over $[2^{17}, 2^{22}]$ on a logarthimic scale, and the width was tuned over $\{64, 128, 256, 512\}$.

\begin{figure}[h!]
    \centering
    \includegraphics[width=0.32\linewidth]{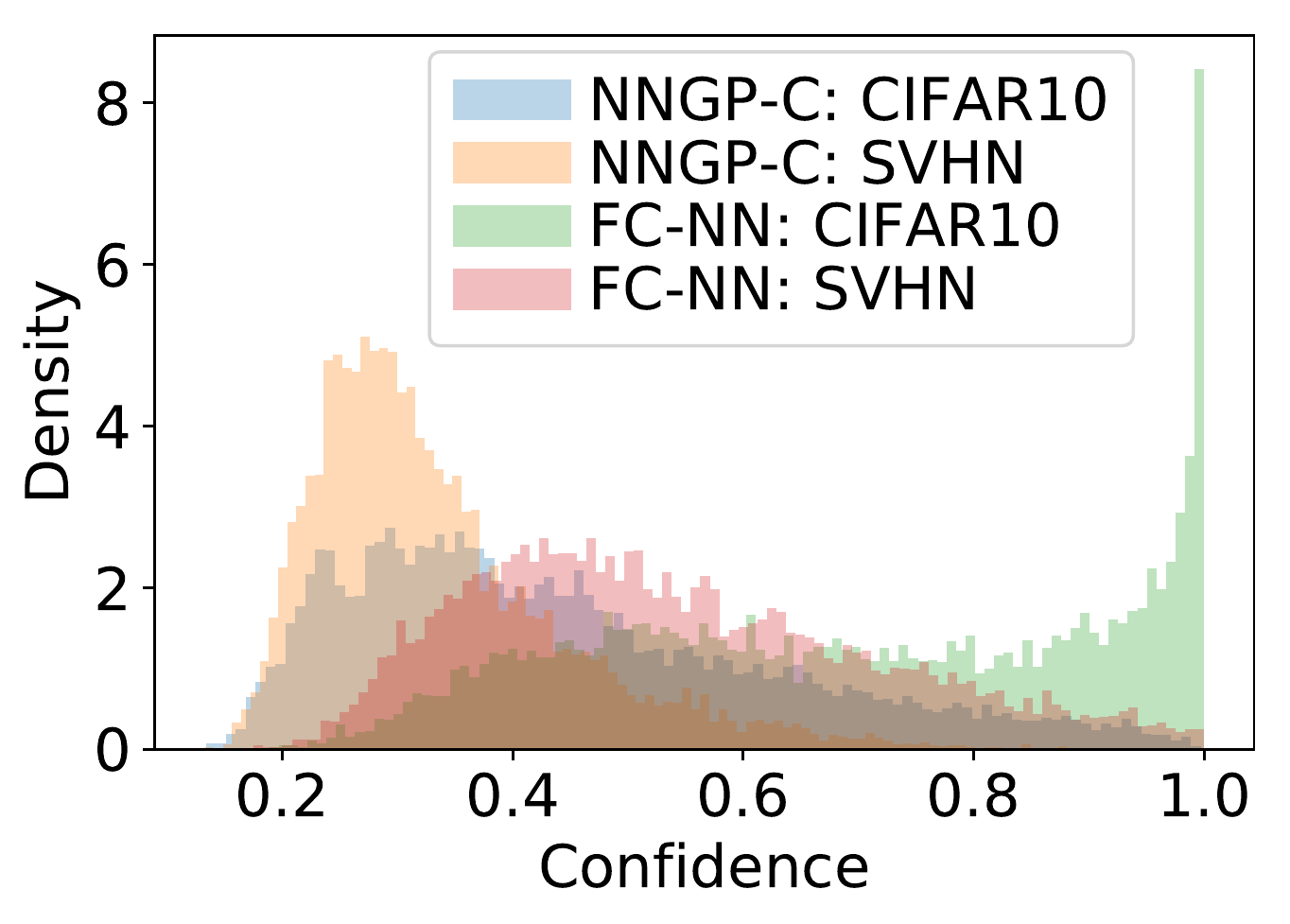}
    \includegraphics[width=0.32\linewidth]{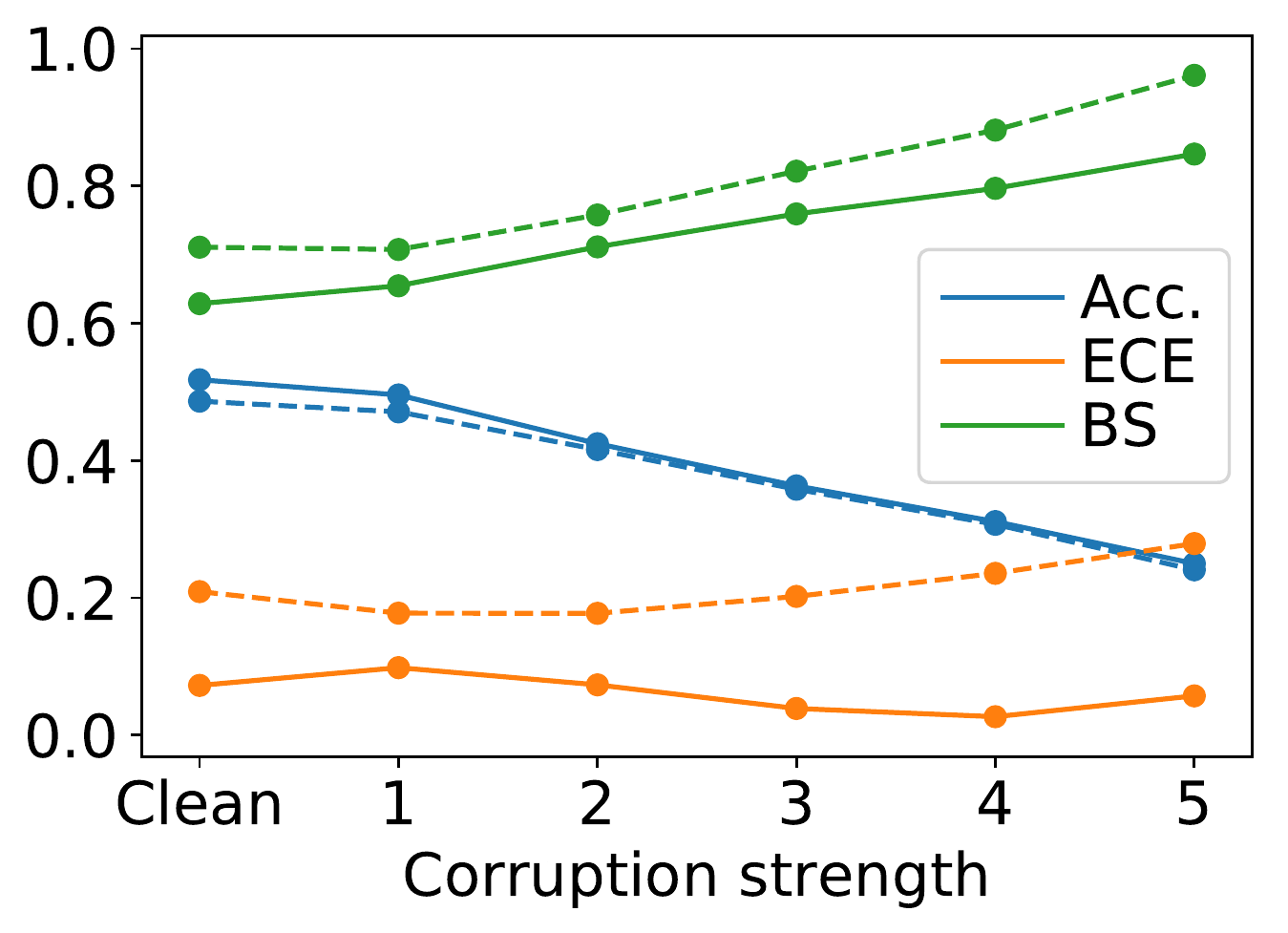}
    \includegraphics[width=0.32\linewidth]{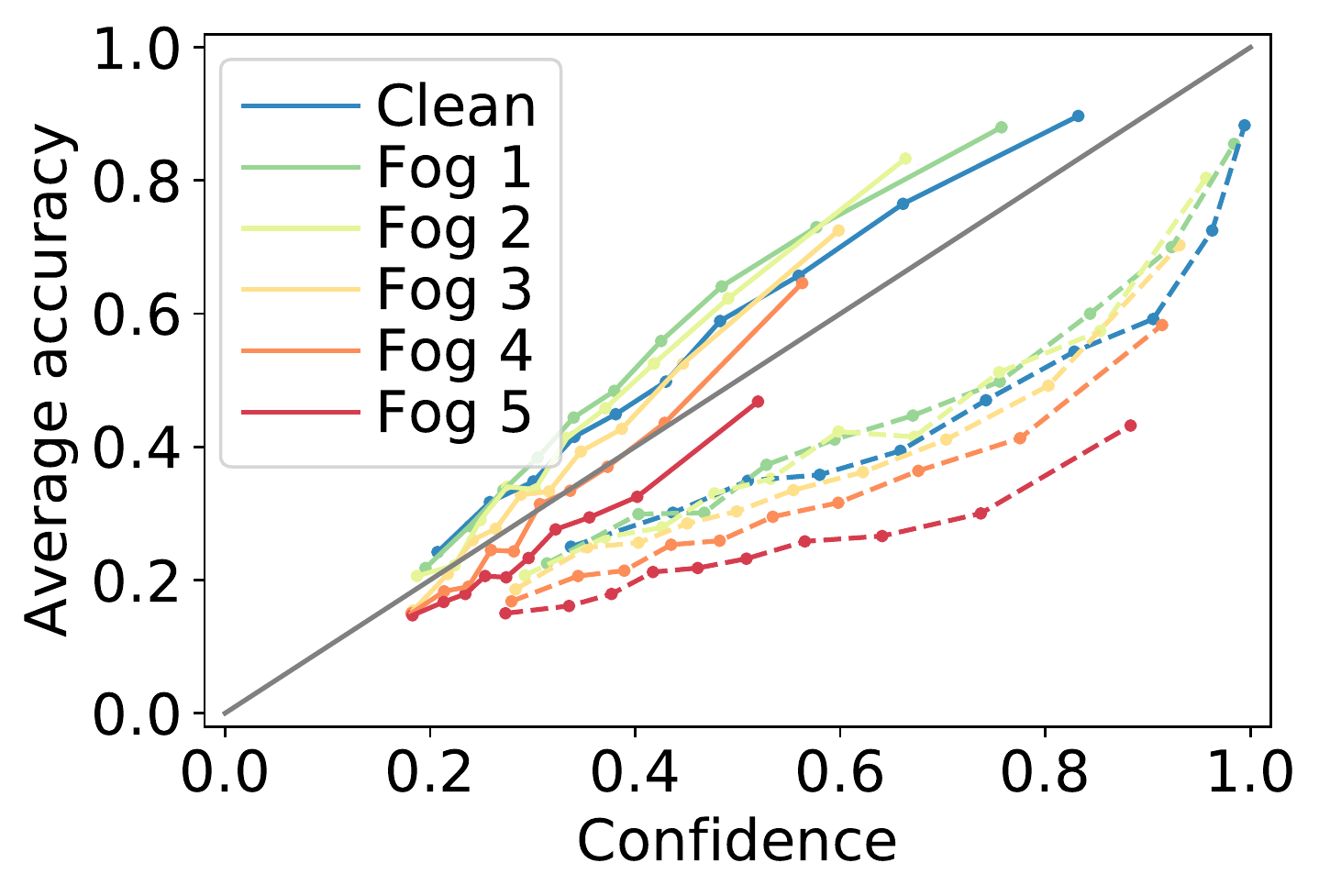} \\
    \includegraphics[width=0.32\linewidth]{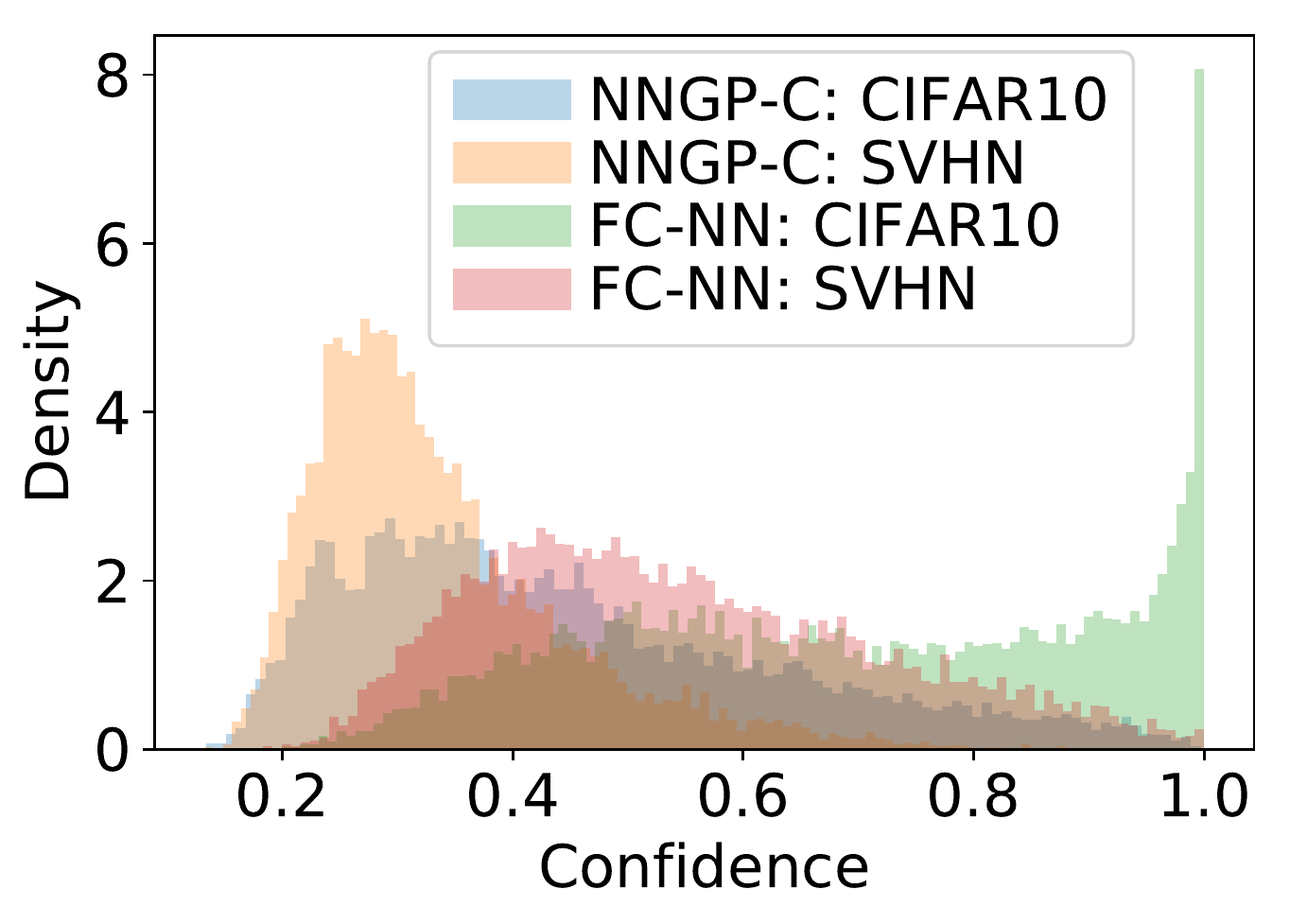}
    \includegraphics[width=0.32\linewidth]{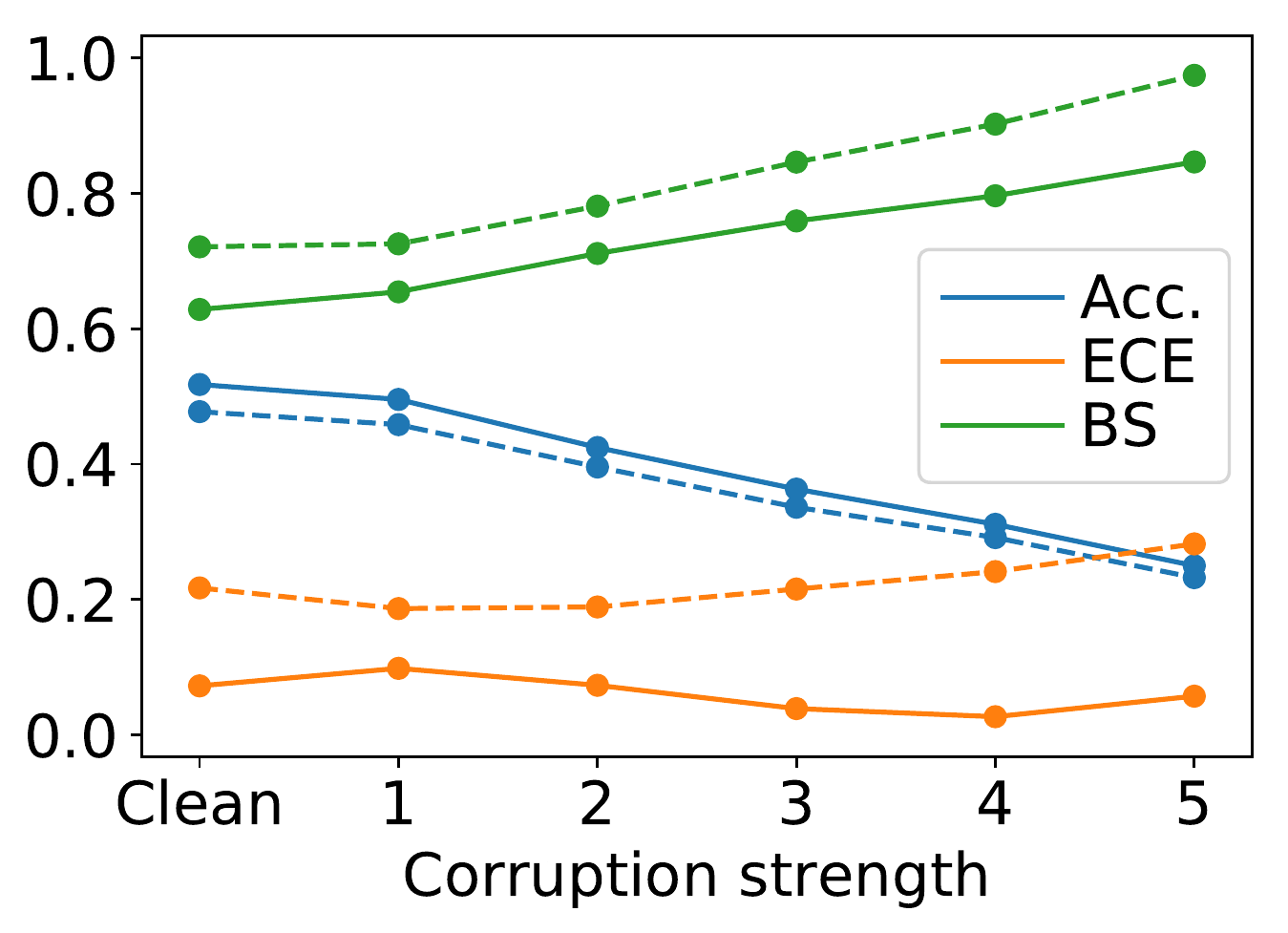}
    \includegraphics[width=0.32\linewidth]{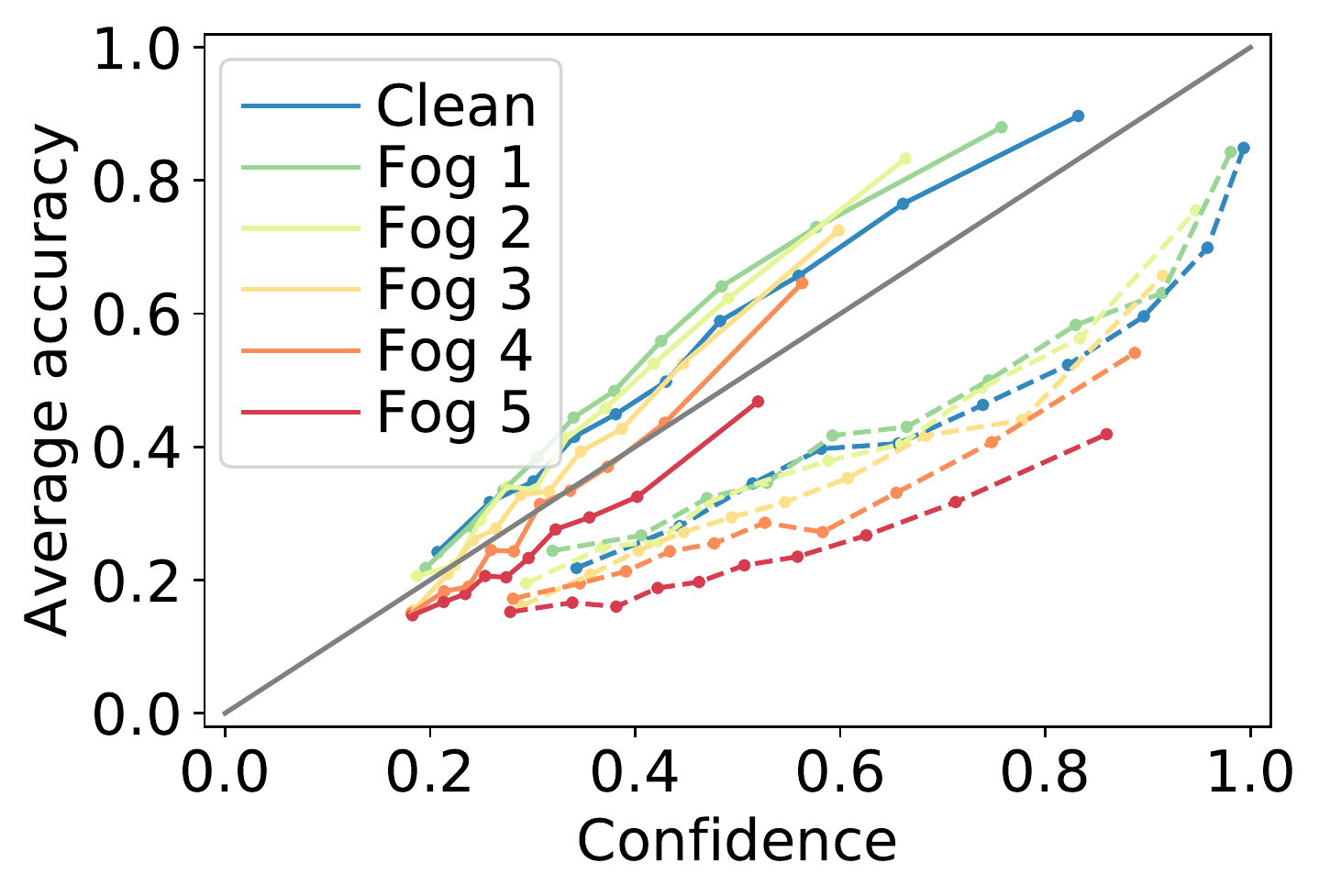} \\
    \includegraphics[width=0.32\linewidth]{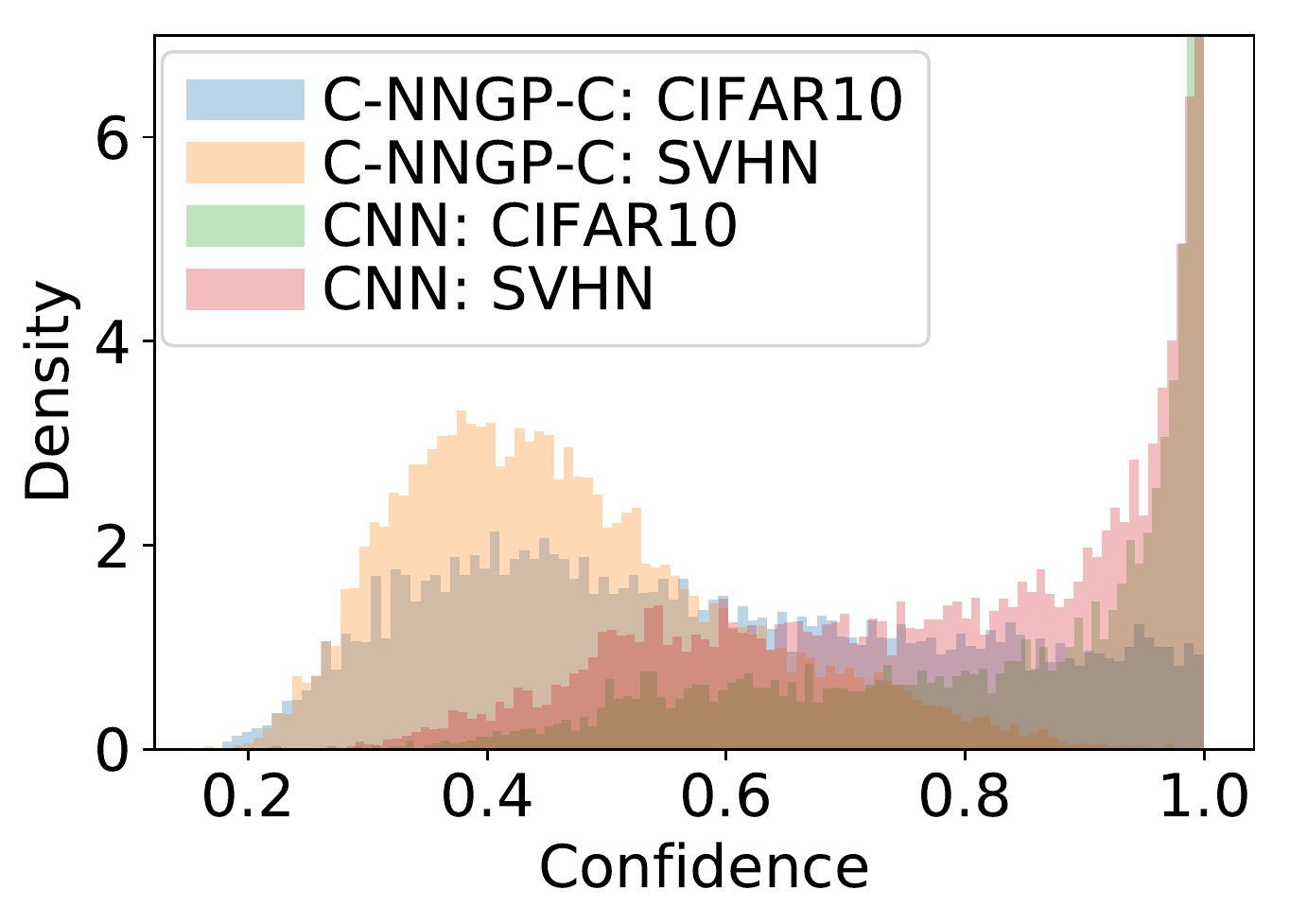}
    \includegraphics[width=0.32\linewidth]{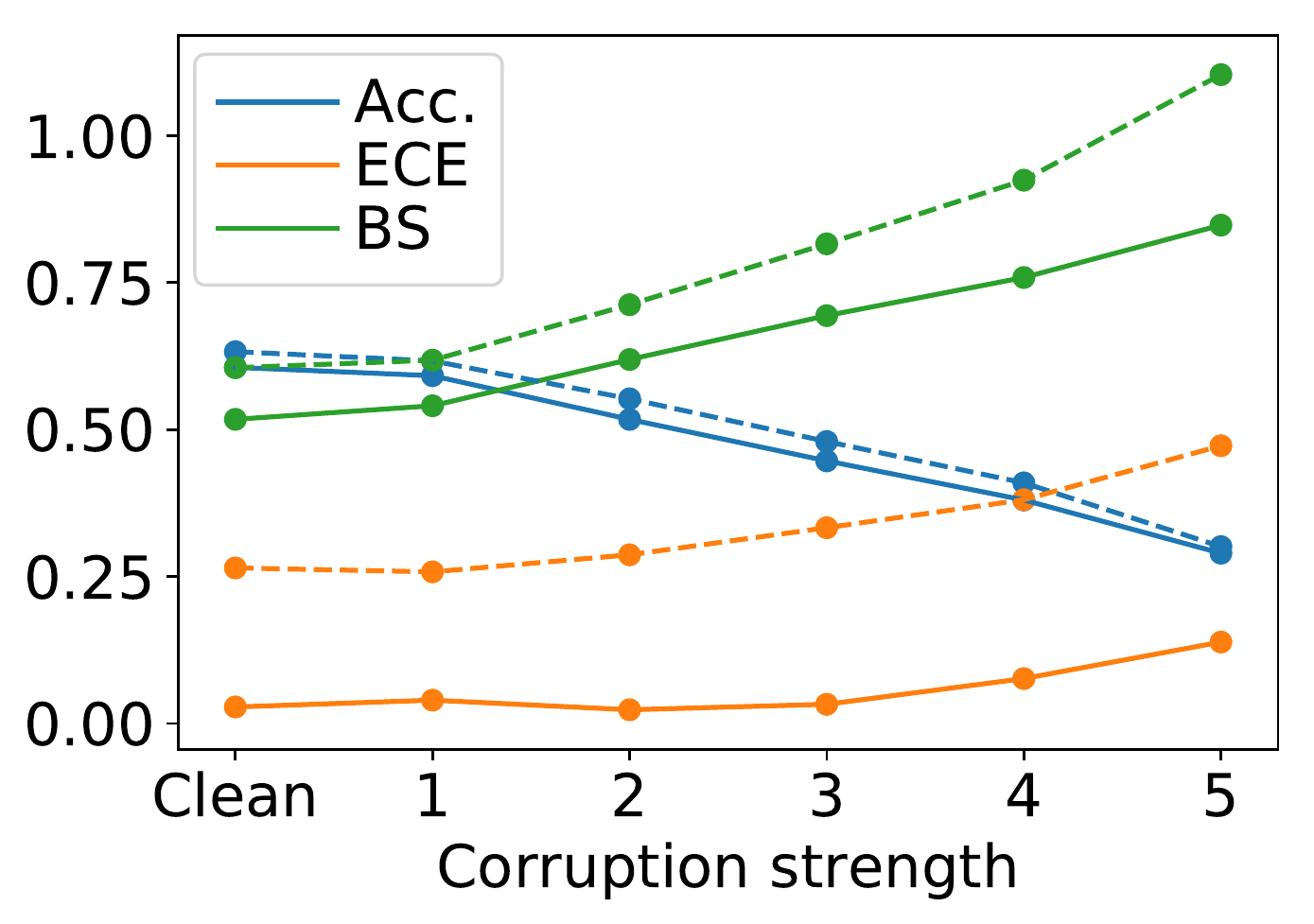}
    \includegraphics[width=0.32\linewidth]{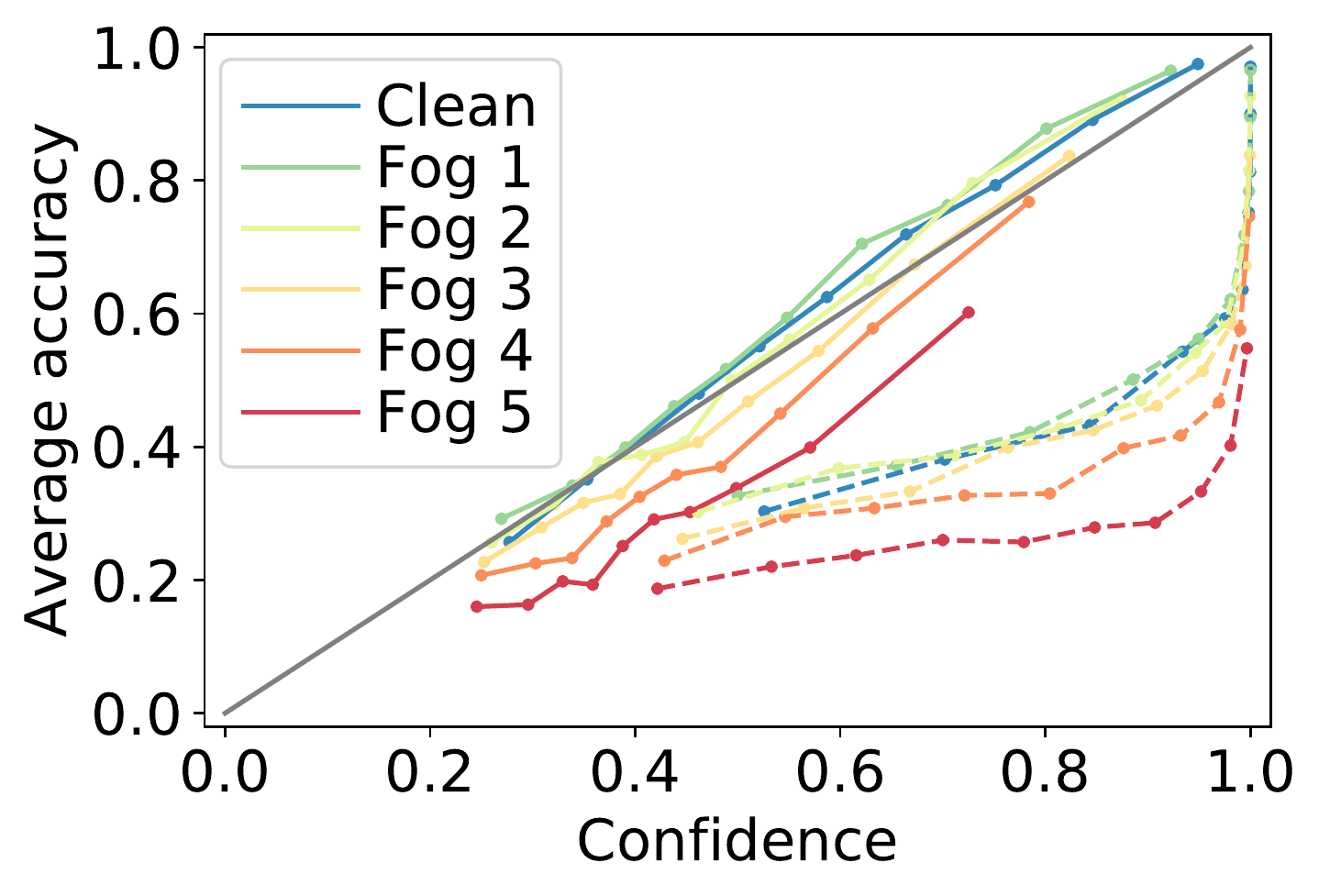} \\
    \caption{Investigating the calibration of Gaussian process classification with NNGP kernels as in Fig.~\ref{fig_GPC}, where we find similar results.
    (\textbf{left column}) Histogram of the confidence of the posterior distribution for each test point. We compare the NNGP-C and a finite width NN on an identically distributed test set (CIFAR10) and an OOD test set (SVHN). (\textbf{middle column}) Performance, NNGP-C is solid and NN dashed, under increasing distributional shift given by the CIFAR10 fog corruption. (\textbf{right column}) We bin the test set into ten bins sorted by confidence, and we plot mean confidence against mean accuracy. (\textbf{top row}) We compare FC-NNGP-C against a FC-NN with the same hyperparameters. (\textbf{middle row}) We compare FC-NNGP-C against a FC-NN, where all of the NN's hyperparameters are tuned independently with Vizier. (\textbf{bottom row}) We compare C-NNGP-C against a CNN, where all of the CNN's hyperparameters are tuned independently with Vizier.}
    \label{fig_sm_GPC}
\end{figure}

\begin{table}[h!]
\centering
\caption{Performance of NNs, where all of the NN's hyperparameters are tuned independently with Google Vizier hyperparameter tuner~\citep{golovin2017}. The NaN entropy measurement is due to the confidence on a specific test point being 1.0 to machine precision. In addition we show results on all Fog corruptions that were omitted from Table~\ref{tab:gpc}.}
\vspace{0.1cm}
\label{tab:sm_gpc}
\renewcommand{\arraystretch}{1.1}% Tighter
\small{
\begin{tabular}{@{}ll|c|c|c|c@{}}
\toprule
Data      &  Metric   & FC-NN (all tuned) & FC-NNGP-C  & CNN (all tuned) & C-NNGP-C  \\
\midrule\midrule
\multirow{4}{*}{CIFAR10}  & ECE & 0.217 & \textbf{0.072} & 0.265 & \textbf{0.031}  \\
                          & Brier Score & 0.721 & \textbf{0.629} & 0.605 & \textbf{0.519} \\
                          & Accuracy & 0.478 & \textbf{0.518} & \textbf{0.632}  & {0.609} \\
                          & NLL & 17967 & \textbf{14007} & 21160 & \textbf{11215} \\
                          \midrule
\multirow{4}{*}{Fog 1}    & ECE & 0.187 & \textbf{0.098} & 0.258 & \textbf{0.042} \\
                          & Brier Score & 0.726 & \textbf{0.655} & 0.618 & \textbf{0.542}  \\
                          & Accuracy & 0.458 & \textbf{0.496}  & \textbf{0.617} & {0.590}  \\
                          & NLL & 17372 & \textbf{14671} & 19484 & \textbf{11755} \\
                          \midrule
\multirow{4}{*}{Fog 2}    & ECE & 0.189 & \textbf{0.073} & 0.287 & \textbf{0.018} \\
                          & Brier Score & 0.781 & \textbf{0.711}  & 0.712 & \textbf{0.620} \\
                          & Accuracy & 0.396 & \textbf{0.425} & \textbf{0.552} & {0.519} \\
                          & NLL & 18930 & \textbf{16233} & 21657 & \textbf{13819} \\
                          \midrule
\multirow{4}{*}{Fog 3}    & ECE & 0.215 & \textbf{0.039}  & 0.333 & \textbf{0.03}  \\
                          & Brier Score & 0.846 & \textbf{0.759} & 0.816 & \textbf{0.694}  \\
                          & Accuracy & 0.336 & \textbf{0.363} & \textbf{0.480} & {0.445} \\
                          & NLL & 21247 & \textbf{17638}  & 25703 & \textbf{16019} \\
                          \midrule
\multirow{4}{*}{Fog 4}    & ECE & 0.241 & \textbf{0.026} & 0.381 & \textbf{0.071}  \\
                          & Brier Score & 0.902 & \textbf{0.797}  & 0.924  & \textbf{0.758}  \\
                          & Accuracy & 0.292 & \textbf{0.311}  & \textbf{0.409} & {0.380}  \\
                          & NLL & 23746 & \textbf{18867}  & 30844 & \textbf{18353}  \\
                          \midrule
\multirow{4}{*}{Fog 5}    & ECE & 0.282 & \textbf{0.057}  & 0.472 & \textbf{0.134} \\
                          & Brier Score & 0.975 & \textbf{0.847}  & 1.104 & \textbf{0.846} \\
                          & Accuracy & 0.232 & \textbf{0.251} & \textbf{0.301} & {0.289}  \\
                          & NLL & 27493 & \textbf{20694}  & 41058 & \textbf{22014}  \\
                          \midrule
\multirow{2}{*}{SVHN}     & Mean Confidence & 0.542 & 0.335 & 0.794 & 0.463 \\
                          & Entropy & 1.208 & 1.840 & 0.524 & 1.403 \\
                          \midrule
\multirow{2}{*}{CIFAR100} & Mean Confidence & 0.654 & 0.398  & 0.847 & 0.474  \\
                          & Entropy & 0.930 & 1.663  & NaN & 1.420 \\
\end{tabular}
}
\end{table}

\clearpage
\begin{figure}[h]
    \centering
    \includegraphics[width=\columnwidth]{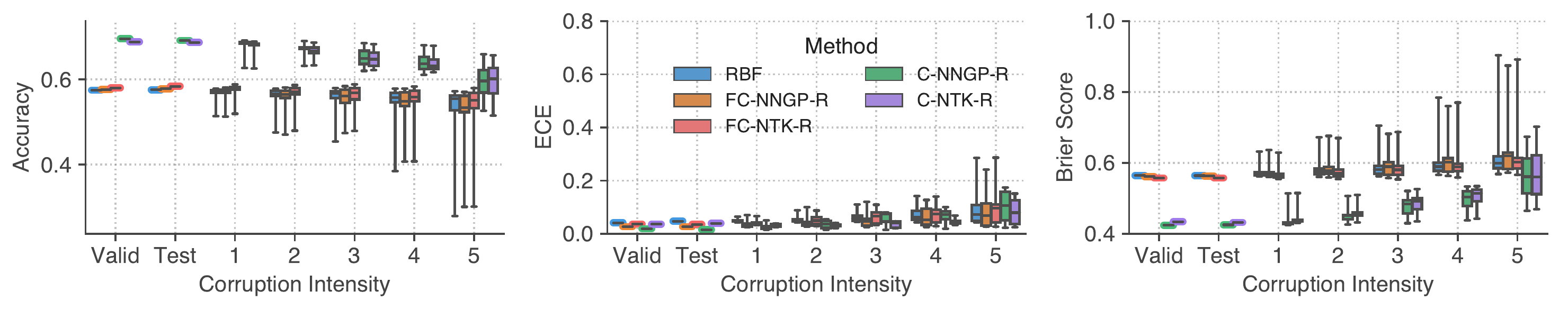}
    \caption{Uncertainty metrics across shift levels on CIFAR10 using NNGP-R and NTK-R. CNN kernels perform best as well as being more robust to corruption. The NTK corresponding to the same architecture shows similar robustness properties.}
    \label{fig:gpr_ntk_box}
\end{figure}

\section{Comparison of Heuristics for Generating Confidences from NNGP-R}

In Secs.~\ref{sec:NN-GPR} and~\ref{sec:gpll}, we utilized a heuristic to generate confidence from exact GPR posterior distribution. Here we denote the heuristic described in Eq.~(\ref{eq:appr-posterior}) as \emph{exact} confidence, which is the probability of a class probit being maximal under an independent multivariate Gaussian distribution. 
We consider two more heuristics. One is denoted \emph{pairwise}, where we take confidence to be proportional to the probability that the $i$-th class probit is larger than other probits in pairwise fashion, \emph{i.e.}
\al{
    p_x(i)\propto \P[z_i > z_j,  \forall j \neq i] = 
     \prod_{j \neq i} p(y_i > y_{j}| x, \X, \Y) =  \prod_{j \neq i} \Phi\left(\frac{\mu_i - \mu_j}{\sqrt{\sigma_i^2 + \sigma_j^2}}\right)\,,
\label{eq:appr-posterior-pair}
}
where $\Phi(\cdot)$ is Gaussian cumulative distribution function. In order to obtain confidence, we normalize by the sum so that the heuristic confidence sums up to $1$. This is following the one-vs-one multiclass classification strategy~\cite{hastie1998classification}. 

We note that, we introduce temperature scaling with temperature $T$ by replacing posterior variances as
\begin{equation}
    \sigma_T^2 = T \sigma^2 \,.
\end{equation}

Another heuristic is denoted \emph{softmax}, where we apply the softmax function to the posterior mean:
\begin{equation}
    p_x(i) \deq \sigma(\mu)_i = \frac{e^{ \mu_i / \sqrt{T}}}{\sum_j e^{ \mu_j / \sqrt{T}}}\,.
\end{equation}
In this case, the posterior variance is not used to construct the heuristic confidences.

\begin{figure}[hb!]
    \centering
    \includegraphics[width=.47\columnwidth]{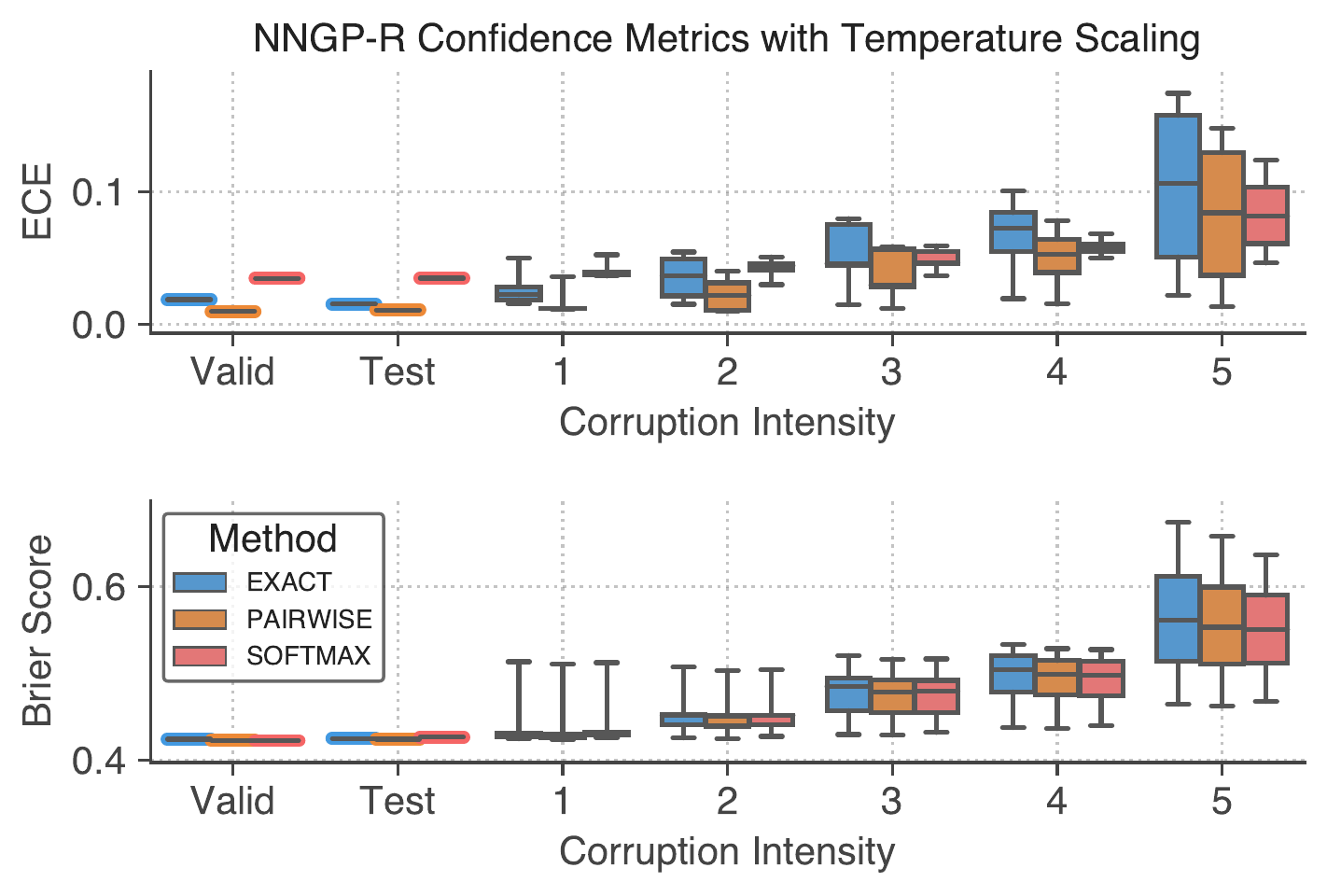}
    \includegraphics[width=.47\columnwidth]{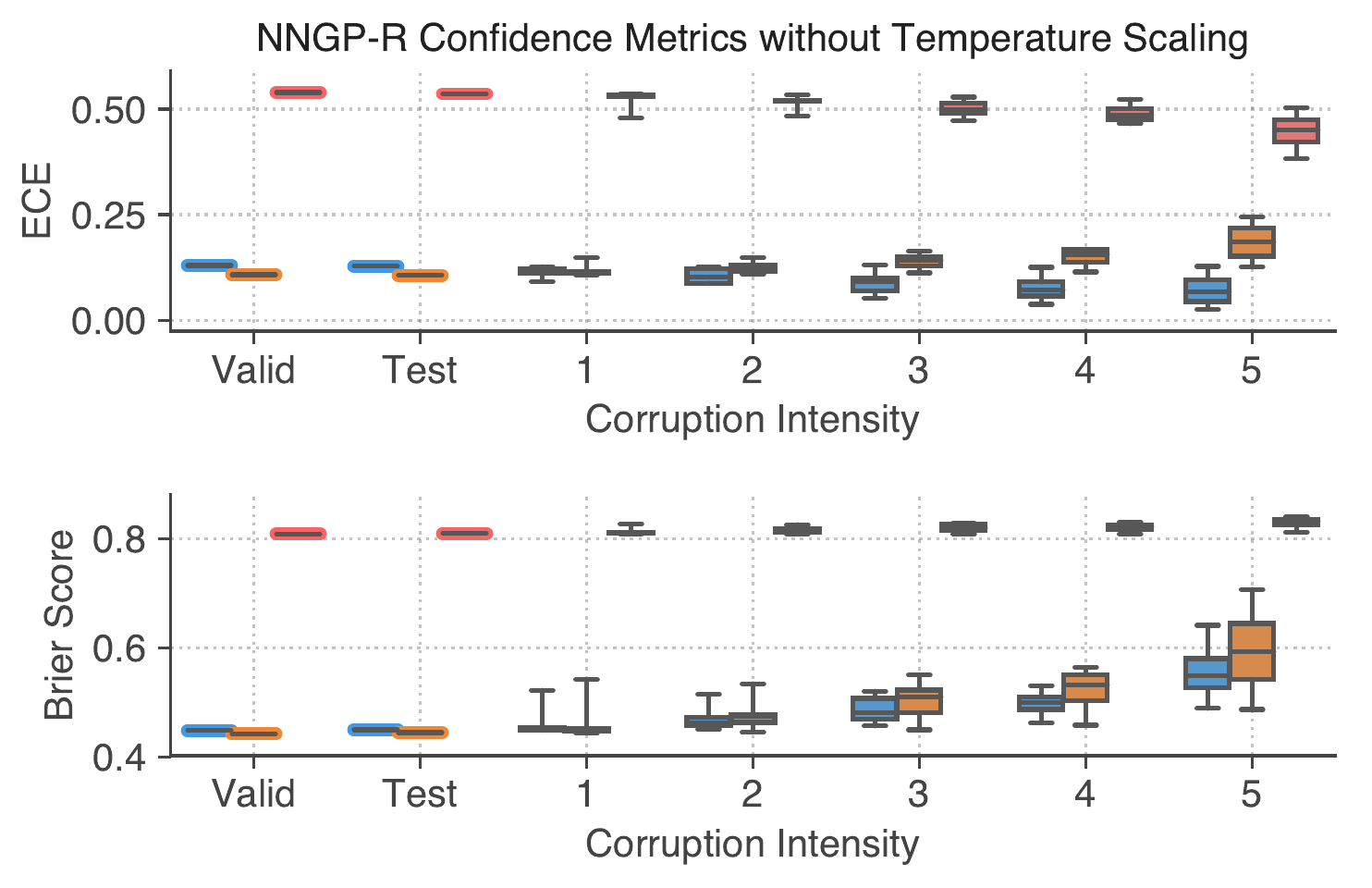}
    \caption{Comparison of NNGP-R based confidence measures on CIFAR10 corruptions using C-NNGP-R. Left column uses temperature scaling based on a validation set whereas right column uses $T=1$. We see that softmax confidence requires adjusting temperature which is equivalent to modifying prior variance to be calibrated whereas exact and pairwise heuristic confidence is calibrated without needing to modify the prior variance.}
    \label{fig:conf-compare}
\end{figure}

A comparison for these three-different heuristics for C-NNGP-R is shown in Fig.~\ref{fig:conf-compare} with and without temperature scaling. We note that \emph{exact} and \emph{pairwise} heuristics remain well calibrated without temperature scaling. However with temperature scaling the \emph{softmax} heuristic can be competitive to other heuristics. In Sec.~\ref{sec:NN-GPR} and~\ref{sec:gpll}, we focused on the \emph{exact} heuristic.

\begin{table}[h!]
\label{tab:gpr}
\centering
\caption{Quartiles of Brier score, negative-log-likelihood and ECE over all CIFAR10 corruptions for methods in Figure~\ref{fig:gpr_box}. }
\vspace{0.2cm}
\begin{tabular}{l|lll}
\toprule
Method/Metric &    RBF & FC-NNGP-R & C-NNGP-R \\
\midrule
Brier Score (25th)  &  0.568 &   0.569 &        \textbf{0.435} \\
Brier Score (50th)  &  0.580 &   0.586 &       \textbf{0.464} \\
Brier Score (75th)  &  0.599 &   0.613 &        \textbf{0.515} \\
\midrule
Gaussian NLL (25th) &  0.147 &   0.612 &        \textbf{0.108} \\
Gaussian NLL (50th) &  \textbf{0.270} &   0.830 &        0.457 \\
Gaussian NLL (75th) & \textbf{0.447} &   1.099 &        1.027 \\
\midrule
NLL (25th)          &  1.331 &   1.351 &        \textbf{1.002} \\
NLL (50th)          &  1.363 &   1.398 &        \textbf{1.079} \\
NLL (75th)          &  1.415 &   1.466 &        \textbf{1.178} \\
\midrule
ECE (25th)          &  0.048 &   0.030 &        \textbf{0.022} \\
ECE (50th)          &  0.052 &   \textbf{0.039} &        0.046 \\
ECE (75th)          &  0.069 &   \textbf{0.065} &        0.071 \\
\midrule
Accuracy (75th)     &  0.573 &   0.574 &        \textbf{0.683} \\
Accuracy (50th)     &  0.566 &   0.561 &        \textbf{0.659} \\
Accuracy (25th)     &  0.549 &   0.541 &        \textbf{0.628} \\
\bottomrule
\end{tabular}
\end{table}

\clearpage
\section{Additional Figures for NNGP-LL}

\begin{figure}[h!]
    \centering
    \includegraphics[width=\columnwidth]{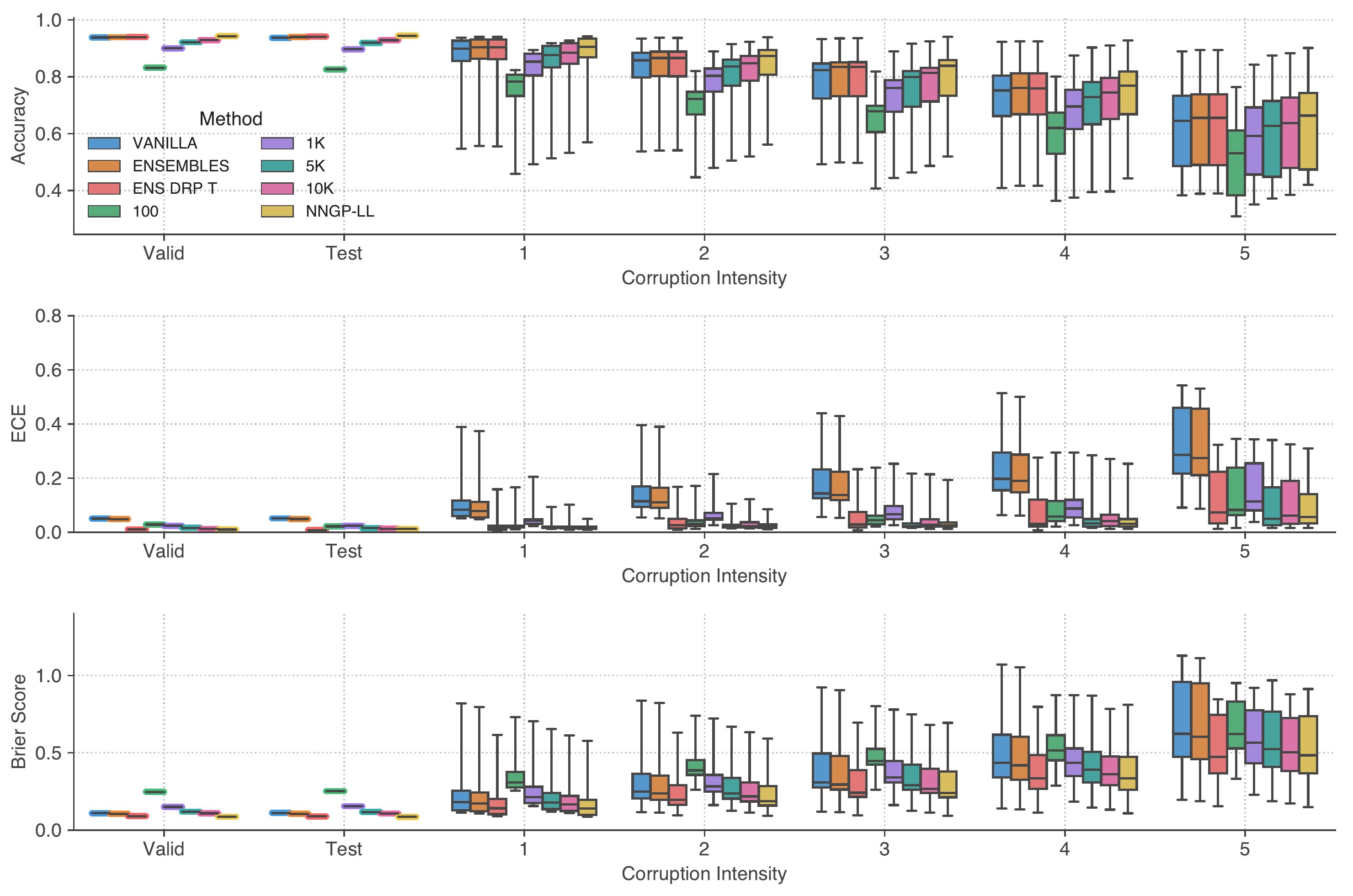}
    \caption{Uncertainty metrics across corruption levels on CIFAR10 using NNGP-LL with EfficientNet-B3 embedding. Baseline NNs are trained on CIFAR10 with parameters of body network fixed. See Table~\ref{tab:gpll_datasize} for quartile comparison and \Figref{fig:gpll} and Table~\ref{tab:gpll} for comparison with fine tuning of all the embedding's weights. \Figref{fig:nngp-vs-rbf} compares use of NNGP and RBF kernel for NNGP-LL settings.
    }
    \label{fig:gpll_datasize}
\end{figure}

\begin{figure}[h!]
    \centering
    \includegraphics[width=\columnwidth]{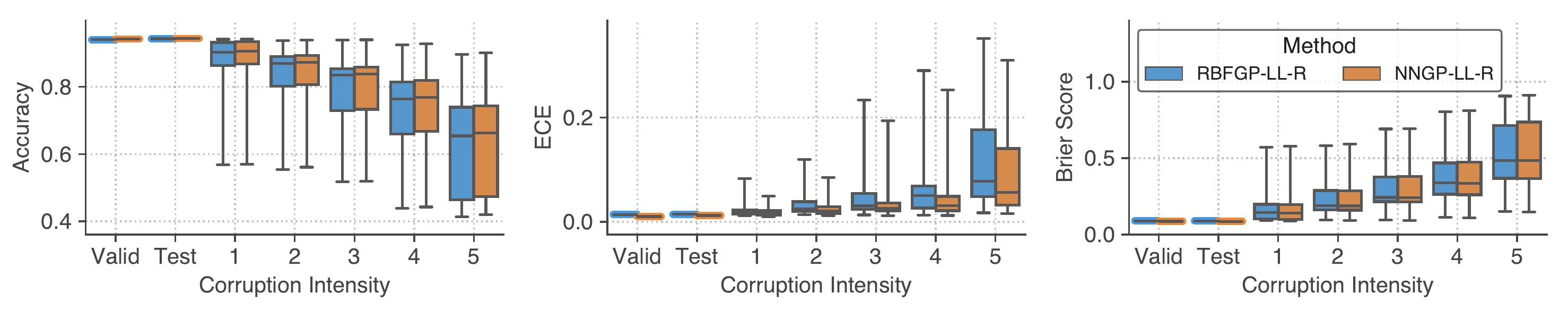}
    \caption{Comparison of NNGP-LL using NNGP head vs RBF GP head  on EfficientNet-B3 embedding. While there are slight advantage of using NNGP head over RBF-GP overall they provide very similar benefits as corruption intensity increase.}
    \label{fig:nngp-vs-rbf}
\end{figure}

The results in the main text focused on a fixed embedding (see Table~\ref{tab:gpll_datasize}) and we show additional results for this as a box-plot in Fig.~\ref{fig:gpll_datasize} here. However, it is also common in practice to tune all of the embedding's weights and simply initialize at their pre-triained values. We explore this setting in the supplement in Table~\ref{tab:gpll} and Fig.~\ref{fig:gpll} by considering the EfficientNet-B3 embedding and fine tuning it on CIFAR10. We also show a results comparing the FC-NNGP-LL with using the standard RBF kernel for the same purpose (see Fig.~\ref{fig:nngp-vs-rbf}).

\begin{table}[h!]
    \centering
\caption{Quartiles of Brier score, negative-log-likelihood, and ECE over all CIFAR10 corruptions for methods in Figure~\ref{fig:gpll} using EfficientNet-B3 embedding. The first five columns are the same as the methods in Table~\ref{tab:gpll_datasize}, whereas the last two columns use the same embedding architecture but where all the weights are fine tuned (FT) on CIFAR10. See Fig.~\ref{fig:gpll} for the corresponding box-plot.}
\label{tab:gpll}    
\resizebox{\textwidth}{!}{%
\begin{tabular}{l|ccccc|cc}
\toprule
{} &     LL & \begin{tabular}{c} LL \\T Scaling\end{tabular} & LL Ens & \begin{tabular}{c} LL Ens \\ Drp T\end{tabular} & NNGP-LL &  FT-NN & FT-NNGP-LL \\
\midrule
Brier Score (25th)  &  0.230 &        0.191 &  0.218 &        0.182 &   0.173 &  0.083 &     \textbf{0.081} \\
Brier Score (50th)  &  0.351 &        0.281 &  0.331 &        0.265 &   0.271 &  0.155 &     \textbf{ 0.153} \\
Brier Score (75th)  &  0.521 &        0.422 &  0.511 &        0.410 &   0.397 &  \textbf{0.341} &      0.361 \\
\midrule
NLL (25th)          &  0.913 &        0.406 &  0.823 &        0.382 &   0.367 &  \textbf{0.166} &      0.196 \\
NLL (50th)          &  1.517 &        0.612 &  1.411 &        0.569 &   0.586 &  \textbf{0.320} &      0.374 \\
NLL (75th)          &  2.662 &        0.988 &  2.492 &        0.932 &   0.905 &  \textbf{0.730} &      0.868 \\
\midrule
ECE (25th)          &  0.104 &        0.016 &  0.098 &        0.016 &   0.017 &  \textbf{0.005} &      0.012 \\
ECE (50th)          &  0.160 &        0.030 &  0.154 &        0.028 &   0.025 &  \textbf{0.015} &      0.022 \\
ECE (75th)          &  0.247 &        0.092 &  0.243 &        0.079 &   0.044 &  0.051 &      \textbf{0.046} \\
\midrule
Accuracy (75th)     &  0.869 &        0.869 &  0.875 &        0.875 &   0.884 &  0.945 &      \textbf{0.947} \\
Accuracy (50th)     &  0.802 &        0.802 &  0.812 &        0.813 &   0.813 &  0.894 &      \textbf{0.896} \\
Accuracy (25th)     &  0.714 &        0.714 &  0.719 &        0.718 &   0.722 &  \textbf{0.758} &      0.748 \\
\bottomrule
\end{tabular}%
}
\end{table}

\begin{figure}[h!]
    \centering
    \includegraphics[width=\columnwidth]{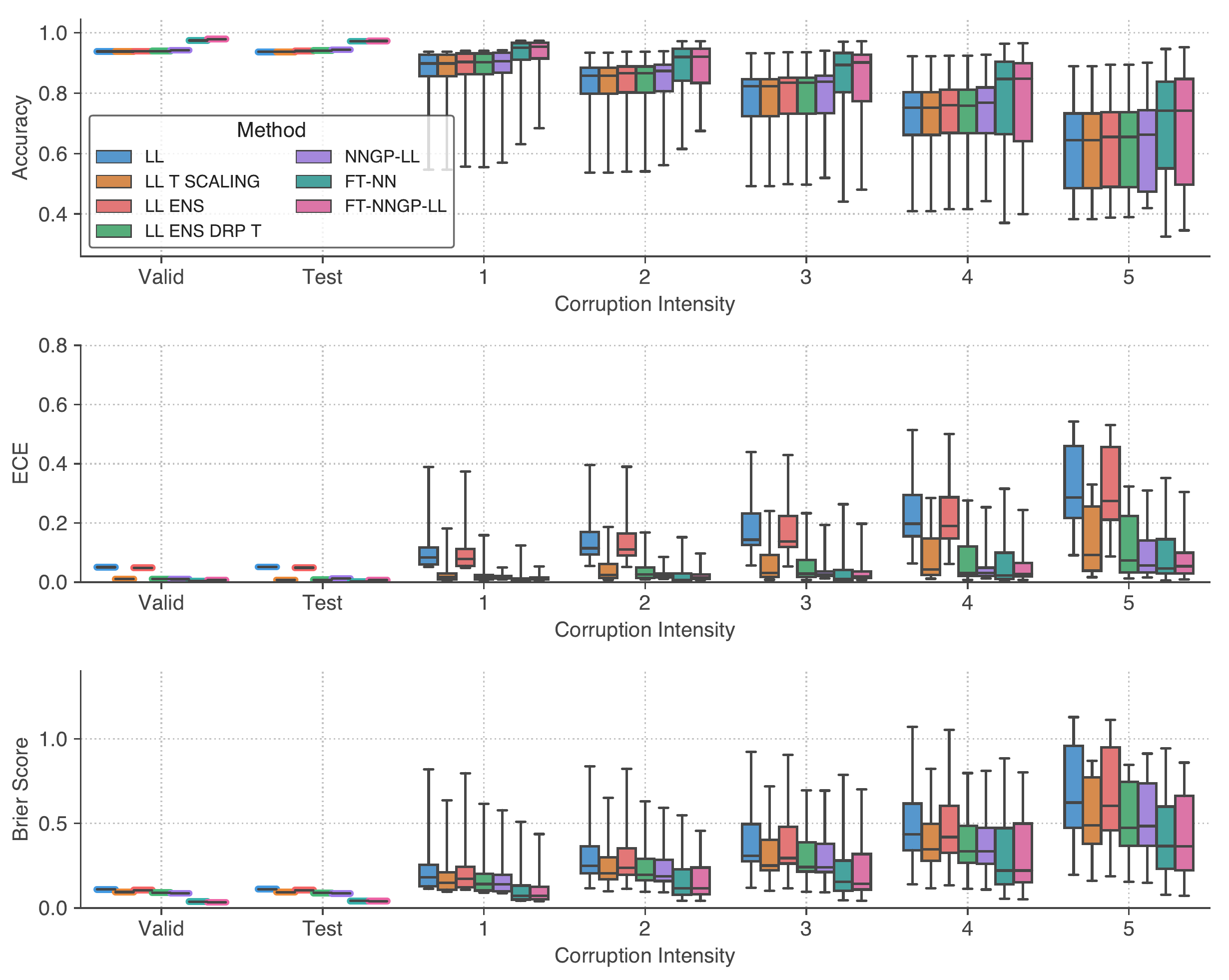}
    \caption{Uncertainty metrics across corruption levels on CIFAR10 using NNGP-LL with EfficientNet-B3 embedding. Baseline NNs are either last layer (LL) trained on CIFAR10 with parameters for body networks fixed or where all the weights are fine tuned (FT). Quartile comparisons can be found in Table~\ref{tab:gpll}.}
    \label{fig:gpll}
\end{figure}

\subsection{A comparison of Ensemble and NNGP-LL on WideResnet}
\begin{figure}[h]
    \centering
    \includegraphics[width=\columnwidth]{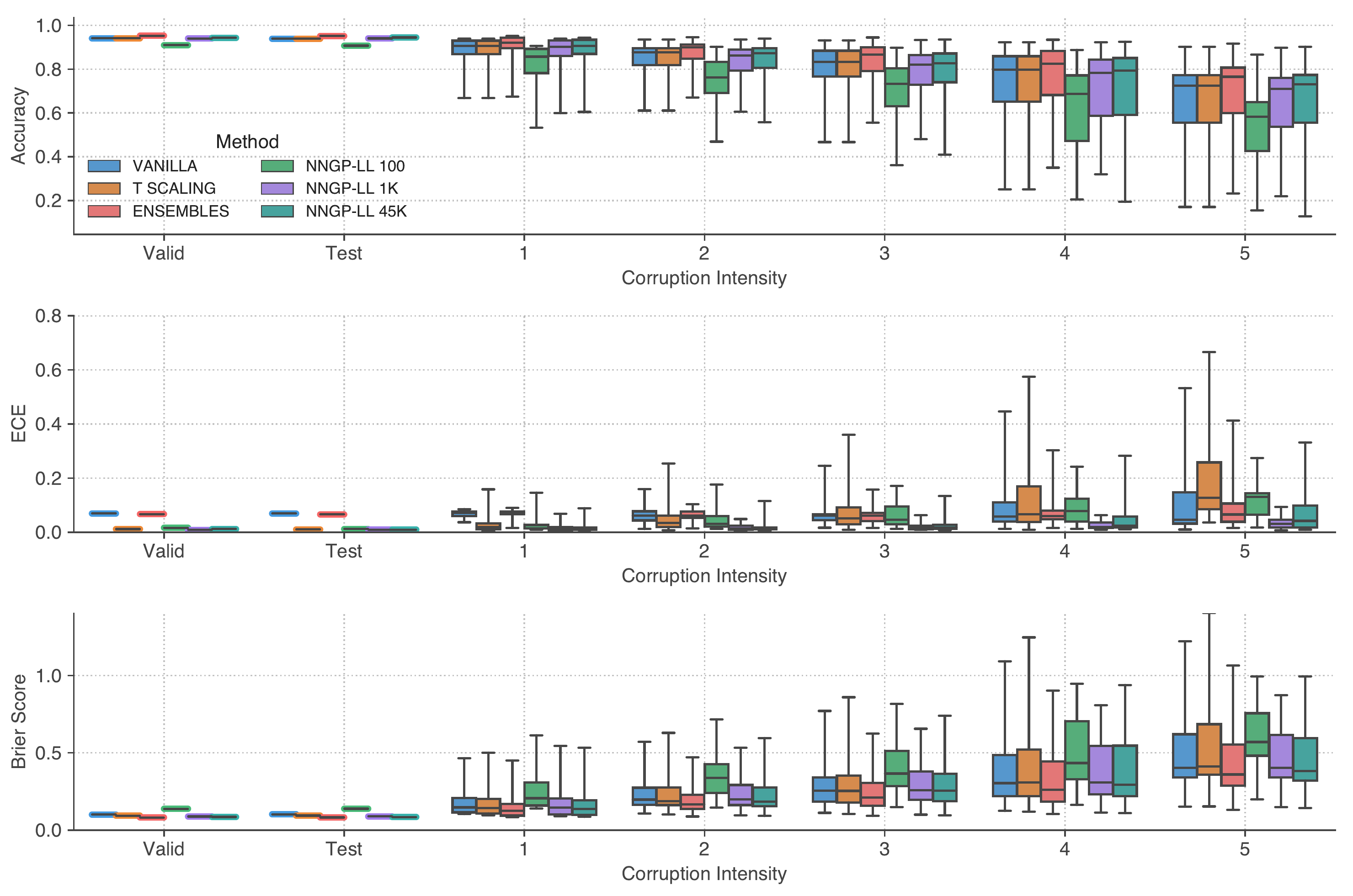}
    \caption{Uncertainty metrics across corruption levels on CIFAR10 using NNGP-LL with MetaInit embedding. Baseline NNs are compared with vanilla training, temperature scaling and ensembles.
    }
    \label{fig:minit_gpll}
\end{figure}

To compare the NNGP-LL method against the gold standard ensemble method, we train a WideResnet 28-10 on CIFAR-10 from scratch with 5 different random initialization. The model is trained using MetaInit \cite{Douphin2019Meta}, Delta-Orthogonal \cite{xiao18a}, mixup \cite{zhang2017mixup} and without BatchNorm \cite{ioffe2015batch}. The model achieves about $94\%$ accuracy on the clean test set. See Table~\ref{tab:gpll_meta} and ~Fig.~\ref{fig:minit_gpll} for the comparison. We find the ECE for NNGP-LL is very competitive, even with small dataset size, compared to baseline methods including ensembles.

\begin{table}[]
    \centering
\caption{Quartiles of Brier score, negative-log-likelihood, ECE and accuray over all CIFAR10 corruptions for MetaInit Embedding NNGP-LL and MetaInit trained networks in Figure~\ref{fig:minit_gpll}.}
\label{tab:gpll_meta}
\resizebox{0.9\textwidth}{!}{%
\begin{tabular}{l|ccc|ccc}
\toprule
Method/Metric & Vanilla & T Scaling & Ensembles & 100 & 1K & NNGP-LL \\
\midrule
Brier Score (25th)  &   0.165 &     0.164 &     \textbf{0.141} &       0.256 &      0.162 &       0.152 \\
Brier Score (50th)  &   0.247 &     0.244 &     \textbf{0.210} &       0.366 &      0.257 &       0.241 \\
Brier Score (75th)  &   0.397 &     0.410 &     \textbf{0.356} &       0.567 &      0.415 &       0.399 \\
\midrule
NLL (25th)          &   0.382 &     0.391 &     \textbf{0.328} &       0.562 &      0.381 &       0.360 \\
NLL (50th)          &   0.553 &     0.576 &     \textbf{0.483} &       0.799 &      0.599 &       0.570 \\
NLL (75th)          &   0.864 &     1.040 &     \textbf{0.781} &       1.214 &      0.967 &       0.925 \\
\midrule
ECE (25th)          &   0.046 &     0.025 &     0.044 &       0.018 &      \textbf{0.011} &       0.012 \\
ECE (50th)          &   0.062 &     0.051 &     0.066 &       0.044 &      \textbf{0.017} &       \textbf{0.017} \\
ECE (75th)          &   0.079 &     0.126 &     0.077 &       0.101 &      \textbf{0.030} &       0.039 \\
\midrule
Accuracy (75th)     &   0.895 &     0.895 &     \textbf{0.916} &       0.824 &      0.889 &       0.896 \\
Accuracy (50th)     &   0.840 &     0.840 &     \textbf{0.866} &       0.738 &      0.821 &       0.834 \\
Accuracy (25th)     &   0.726 &     0.726 &     \textbf{0.761} &       0.586 &      0.703 &       0.716 \\
\bottomrule
\end{tabular}
}
\end{table}
\begin{figure}[h]
    \centering
    \includegraphics[width=\columnwidth]{figs/eb3_gpll_vary_embedding.pdf}    
    \includegraphics[width=\columnwidth]{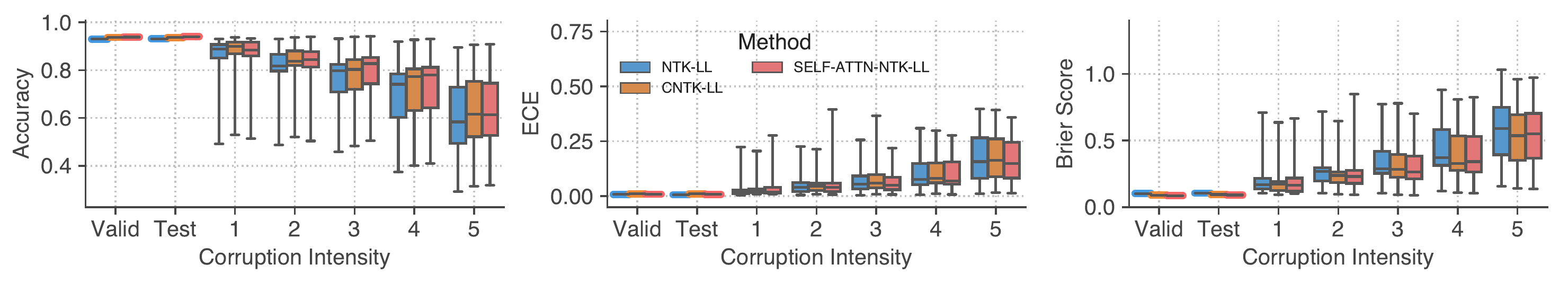} 
    \caption{Comparison between NNGP-LL and NTK-LL with varying embeddings.}
    \label{fig:gpll_embedding_ntk_box}
\end{figure}
\end{document}